%% file: neurips_2024.tex
\documentclass[table]{style/style}

\usepackage{microtype}
\usepackage{hyperref}
\usepackage{url}
\usepackage{booktabs}
\usepackage{graphicx}
\usepackage{lineno}
\usepackage{enumitem}
\usepackage{listings} %
\usepackage{svg}

\definecolor{ darkblue}{rgb}{0, 0, 0.5}
\hypersetup{colorlinks=true, citecolor=darkblue, linkcolor=darkblue, urlcolor=darkblue}

\usepackage{amssymb}
\usepackage{multirow}
\usepackage{bigdelim}
\usepackage{todonotes}
\usepackage{longtable}
\usepackage{tabularray}
\usepackage{wrapfig}
\usepackage[most]{tcolorbox} 
\usepackage{url}
\usepackage{xspace}
\usepackage{svg}
\usepackage[absolute]{textpos} %
\usepackage{algorithm}
\usepackage{algpseudocode}
\algrenewcommand\algorithmicrequire{\textbf{Require:}}
\algrenewcommand\algorithmicensure{\textbf{Ensure:}}
\usepackage{fdsymbol}   %

\usepackage{tabularx}

\usepackage[utf8]{inputenc} %
\usepackage[T1]{fontenc}    %
\usepackage{url}            %
\usepackage{booktabs}       %
\usepackage{amsfonts}       %
\usepackage{nicefrac}       %
\usepackage{microtype}      %
\usepackage[table,prologue,dvipsnames]{xcolor}
\colorlet{lightgray}{White!30!lightgray}
\colorlet{lightblue}{White!70!MidnightBlue}
\usepackage{amsmath}
\usepackage[most]{tcolorbox}
\usepackage{csquotes}
\usepackage{wrapfig}
\usepackage{siunitx}
\usepackage{graphicx}
\usepackage{arydshln}
\usepackage{wrapfig}
\usepackage{enumitem}
\usepackage{soul} %

\usepackage{multirow}
\usepackage{xspace}
\usepackage{adjustbox}
\usepackage{pifont}
\usepackage{caption}
\usepackage{makecell}
\usepackage{subcaption}
\usepackage{bold-extra}
\usepackage{url}
\usepackage{float}
\usepackage{pgf-pie}

\usepackage{hyperref}
\crefname{figure}{Fig.}{Figs.}
\crefname{equation}{Eqn.}{Eqns.}
\crefname{appendix}{Appx.}{Appx.}
\crefname{table}{Table}{Tables}
\crefname{algorithm}{Algorithm.}{Algorithms.}
\crefname{section}{\S}{\S\S}

\definecolor{promptshade}{RGB}{245,245,248}
\colorlet{shadecolor}{promptshade}

\newcommand{\ours}{{RECON}\xspace}

\usepackage{arydshln}
\makeatletter
\def\adl@drawiv#1#2#3{%
        \hskip.5\tabcolsep
        \xleaders#3{#2.5\@tempdimb #1{1}#2.5\@tempdimb}%
                #2\z@ plus1fil minus1fil\relax
        \hskip.5\tabcolsep}
\newcommand{\cdashlineCustom}[1]{%
  \noalign{\vskip\aboverulesep
           \global\let\@dashdrawstore\adl@draw
           \global\let\adl@draw\adl@drawiv}
  \cdashline{#1}
  \noalign{\global\let\adl@draw\@dashdrawstore
           \vskip\belowrulesep}}
\makeatother

\definecolor{linkcolor}{RGB}{0, 0, 128}
\hypersetup{
     colorlinks   = true,
     citecolor    = linkcolor,
     linkcolor    = linkcolor,
     urlcolor     = linkcolor,
}
\usepackage{pifont}%
\usepackage{listings}

\setlist[itemize]{leftmargin=*,itemsep=0em,parsep=0.3em,topsep=0.3em}

\DeclareUnicodeCharacter{2212}{\ensuremath{-}}

\usepackage{tikz}

\usepackage{setspace}

\usepackage{nicematrix}
\newcolumntype{L}[1]{>{\raggedright\let\newline\\\arraybackslash\hspace{0pt}}m{#1}}
\newcolumntype{C}[1]{>{\centering\let\newline\\\arraybackslash\hspace{0pt}}m{#1}}
\newcolumntype{R}[1]{>{\raggedleft\let\newline\\\arraybackslash\hspace{0pt}}m{#1}}
\newcolumntype{P}[1]{>{\centering\let\newline\\\arraybackslash\hspace{0pt}}m{#1}}
\newcommand{\huggingface}{\raisebox{-1.5pt}{\includegraphics[height=1.05em]{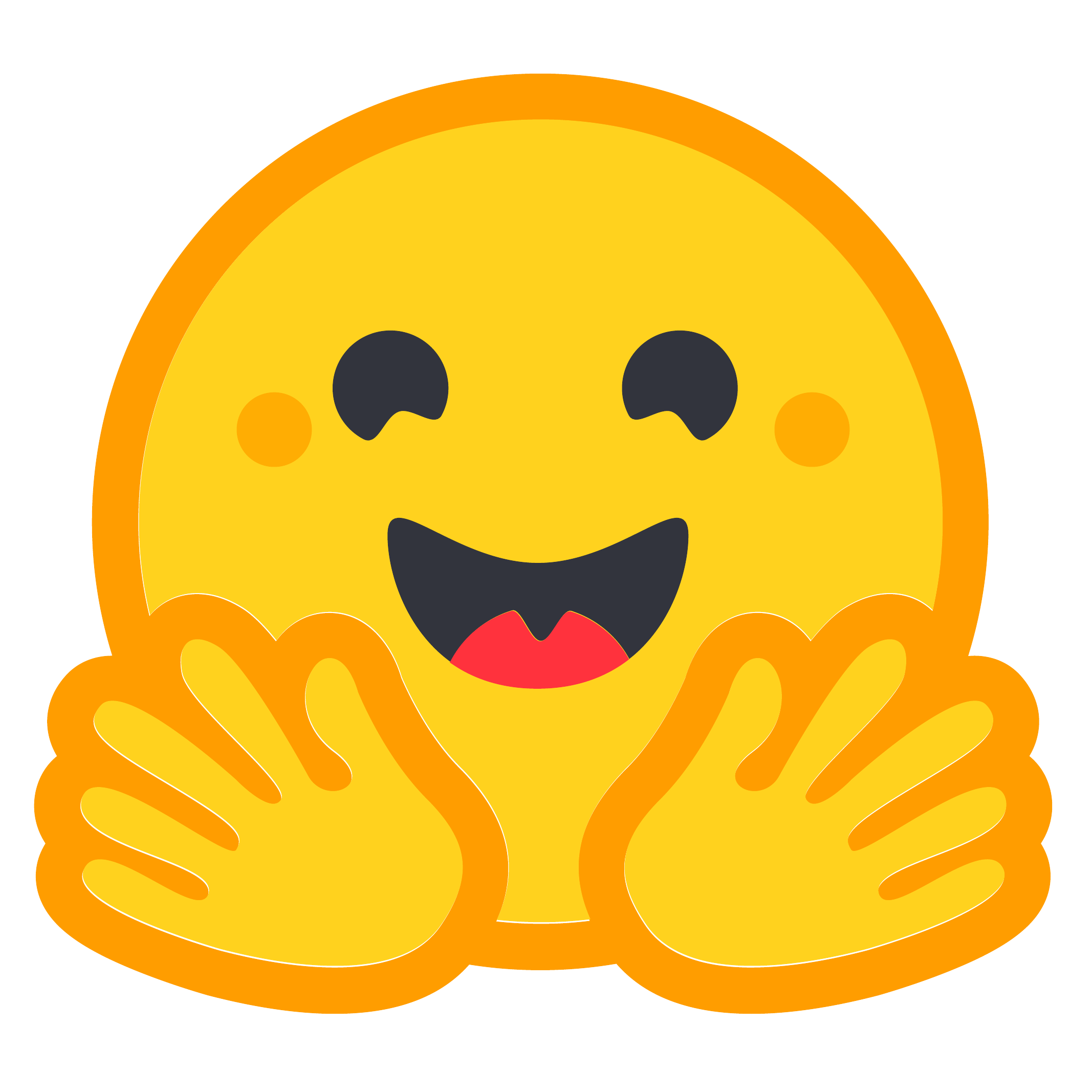}}\xspace}

\newcommand{\github}{\raisebox{-1.5pt}{\includegraphics[height=1.05em]{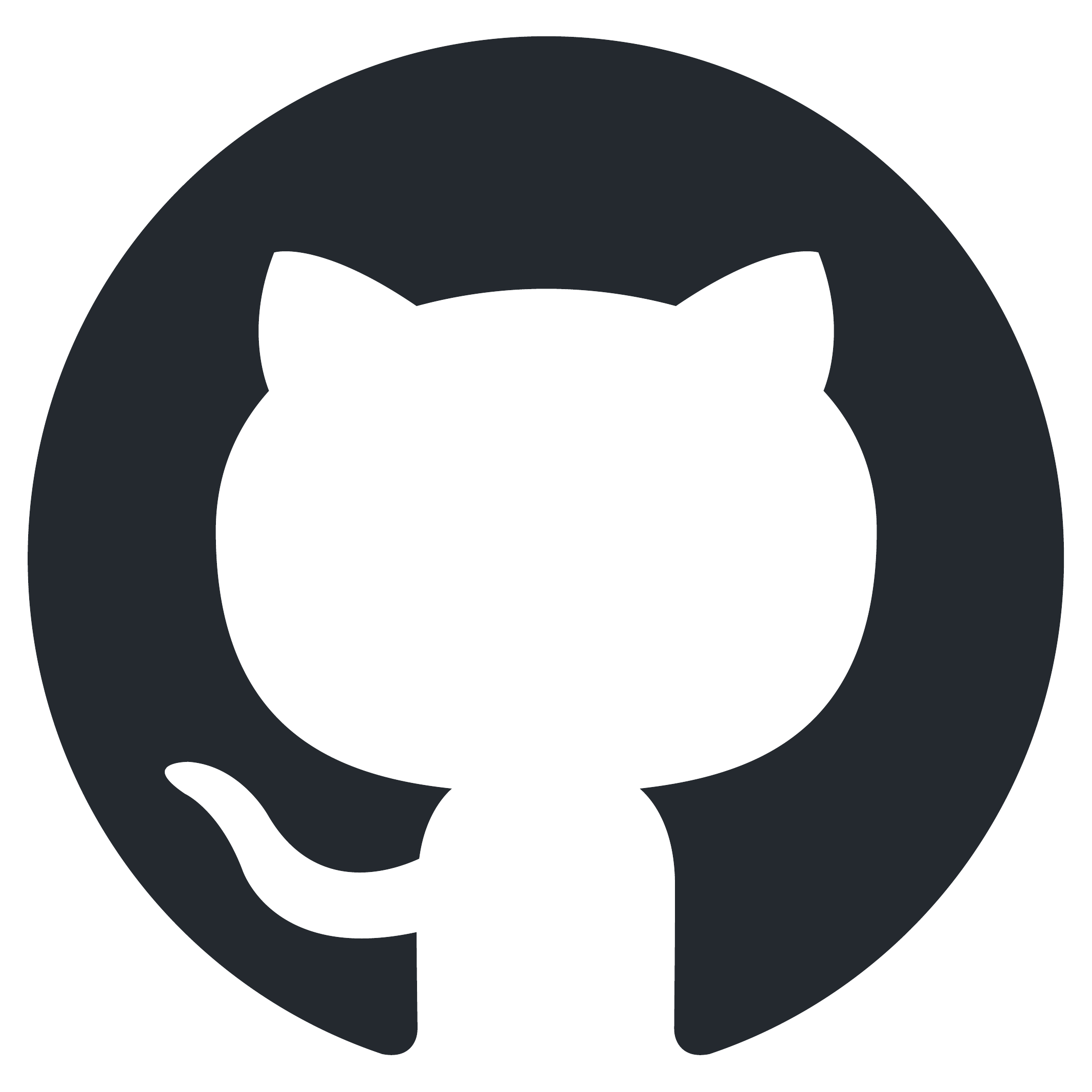}}\xspace}

\title{\LARGE RECON: Reasoning with Condensation for Efficient Retrieval-Augmented Generation}

\authorOne[*,1]{Zhichao Xu}
\authorOne[*,2]{Minheng Wang}
\authorOne[3]{Yawei Wang}
\authorOne[4]{Wenqian Ye}
\authorOne[5]{Yuntao Du}
\authorOne[6]{Yunpu Ma}
\authorOne[7]{\newline Yijun Tian}

\affiliation[1]{University of Utah}
\affiliation[2]{University of Washington}
\affiliation[3]{George Washington University}
\affiliation[4]{University of Virginia}
\affiliation[5]{Shandong University}
\affiliation[6]{Ludwig Maximilian University of Munich}
\affiliation[7]{University of Notre Dame}
\contribution[*]{Equal contribution.}
\contribution[]{\texttt{zhichao.xu@utah.edu} \quad \texttt{meetyijun@gmail.com}}

\abstract{
  Search agents trained with reinforcement learning (RL) interleave reasoning with tool calls in a multi-turn, tool-integrated reasoning (TIR) loop, where each tool invocation returns an environment observation that is appended to the agent's context. As the rollout proceeds, these raw observations accumulate, inflating token cost and diluting the signal available for downstream reasoning. Unlike single-pass retrieve-then-read pipelines, where context compression is a one-time postprocessing step, the multi-turn RL setting requires compression that runs at every observation step while remaining decoupled from policy optimization. We introduce \ours (REasoning with CONdensation), a framework that addresses this challenge by inserting a dedicated observation compressor into the reasoning loop. The compressor is trained via a two-stage curriculum: relevance pretraining on QA datasets followed by multi-aspect distillation from proprietary LLMs, and remains frozen during RL training to preserve policy stability. Integrated into the Search-R1 search-agent pipeline, \ours reduces total context length by 35\%, improves training speed by 5.4\% and inference latency by 30.9\%, while boosting average exact-match by 14.5\% on the 3B agent and 3.0\% on the 7B agent, with particular strength in multi-hop QA. These results establish learned observation compression as a key component for building practical, scalable RL-trained search agents.
}
\setmaintable{
\begin{table}[!h]
    \centering
    \begin{tabular}{l}
        \multicolumn{1}{l}{
            \github~{\sans{Code}}~
            \href{https://github.com/minhengwang/RECON}{\texttt{RECON}}
        } \\
        \multicolumn{1}{l}{
            \huggingface~{\sans{Data}}~
            \href{https://huggingface.co/datasets/allfornancy/msmarco-sft}{\texttt{MS MARCO Relevance Pretraining}}
        } \\
        \multicolumn{1}{l}{
            \huggingface~{\sans{Data}}~
            \href{https://huggingface.co/datasets/allfornancy/mixed-sft-data-alpaca-complete}{\texttt{Summarization Finetuning}}
        }
    \end{tabular}
\end{table}
}

\begin{document}

\maketitle

\section{Introduction}
\label{sec:intro}

Reinforcement learning~\citep[RL,][]{sutton1998reinforcement} has recently emerged as a powerful approach for improving the reasoning capabilities of large language models~\citep[LLMs,][]{jaech2024openaio1systemcard,guo2025deepseek}. A particularly active line of work trains LLMs to act as \emph{search agents}---models that interleave reasoning with calls to external tools (e.g., a retriever or a web search API), reading the returned \emph{environment observations} to inform the next reasoning step. This tool-integrated reasoning (TIR) paradigm extends earlier prompting frameworks such as ReAct~\citep{yao2023react} and Toolformer~\citep{schick2023toolformer} by training the agent end-to-end with reasoning-aware RL, and has yielded methods such as ReSearch~\citep{chen2025learningtoreason}, R1-Searcher~\citep{song2025r1searcher}, DeepResearcher~\citep{zheng2025deepresearcher}, and Search-R1~\citep{jin2025searchr1,jin2025searchr1extension}. These search agents substantially improve performance on knowledge-intensive benchmarks that require chaining and reconciling evidence across multiple sources rather than relying solely on parametric memory~\citep{mallen-etal-2023-trust,asai2024self}, and they generalize the retrieve-then-read framing of classical retrieval-augmented generation~\citep[RAG,][]{Patrick2020rag,trivedi2023ircot} into a true multi-turn agent loop~\citep{lin2025comprehensivesurveyreinforcementlearningbased}.

A defining feature of this loop is that environment observations (i.e., raw retrieved passages, web pages, tool outputs) are appended verbatim to the agent's context after every tool call. In knowledge-intensive settings these observations are typically verbose, redundant, or only loosely relevant to the query. Passing them in unchanged inflates context length, drives up token cost~\citep{jin2025searchr1}, and can even degrade accuracy~\citep{sun2025dynamicrag,chang-etal-2025-main}. The damage is twofold: (1) \emph{computational}---longer contexts increase token usage, latency, and serving costs; and (2) \emph{cognitive}---diluted or noisy observations weaken downstream reasoning~\citep{liu-etal-2024-lost,levy2025moredocuments} and increase hallucination risk~\citep{Huang2025SurveyOnHallucination}. Because the loop is multi-turn, these effects \emph{compound}: each tool call adds another block of raw observation, so context grows roughly linearly in the number of turns, posing memory and throughput challenges in both RL training and deployment~\citep{cao2025skyrl,feng2025groupingroup}.

\input{figures/pipeline}

Compressing observations is a well-studied problem in single-pass RAG~\citep{xu2024recomp,jiang-etal-2023-llmlingua,li-etal-2023-compressing,wang-etal-2024-learning-to-filter}, but the multi-turn search-agent setting introduces design constraints that prior work does not satisfy. Specifically, a compressor for an RL-trained search agent must (i) operate at \emph{every} tool call rather than as a one-time postprocessing pass, so that the observation stream does not regrow turn over turn; (ii) remain \emph{frozen and fully decoupled} from the agent's policy, because jointly optimizing the compressor with the RL objective destabilizes PPO training; and (iii) emit \emph{abstractive, human-readable} observations so that the agent's interleaved reasoning trace stays auditable~\citep{Lanham2023MeasuringFI,baker2025monitoring,korbak2025chain,xu2026beyondcorrectness}. Existing single-pass extractive or hard-prompt-pruning methods~\citep{jiang-etal-2023-llmlingua,li-etal-2023-compressing,wang-etal-2024-learning-to-filter,xu2024recomp} were built for a different regime---one tool call, no policy gradient---and do not directly transfer.

We instantiate this framework within the Search-R1 search-agent pipeline and call the resulting method \textbf{RE}asoning with \textbf{CON}densation (\textbf{\ours}). The compressor is trained in two stages: first, relevance pretraining on MS MARCO~\citep{bajaj2016msmarco} to discriminate useful from irrelevant retrieved passages; then, distillation of multi-aspect summaries from GPT-4o-mini~\citep{hurst2024gpt4osystemcard}, aligning the model with human-preferred properties such as factuality and clarity~\citep{pagnoni-etal-2021-understandingfactuality,kryscinski2020evaluatingthefactualconsistency,liu-etal-2023-improvingsummarizationfactual,ryu-etal-2024-multidimensional}. The compressor remains frozen during RL training to preserve policy stability.

\paragraph{Contributions.} We make the following contributions:
\begin{itemize}
    \item We articulate the design constraints (per-tool-call, frozen-during-RL, abstractive) that distinguish observation compression for RL-trained search agents from compression in single-pass RAG, and show that satisfying them is non-trivial: in particular, removing the relevance-pretraining stage causes the 7B agent to collapse during RL training.
    \item We instantiate the framework as \ours and integrate it into Search-R1 as a plug-and-play module sitting between the search tool and the agent. The two-stage curriculum---relevance pretraining followed by multi-aspect distillation---yields a lightweight 3B compressor that is reusable across summarization aspects (clarity, factuality, completeness, $\ldots$) without retraining.
    \item Across seven QA benchmarks, \ours improves average exact-match by 14.5\% on the 3B agent and 3.0\% on the 7B agent over Search-R1, with the largest gains on multi-hop QA where evidence must be reconciled across many tool calls. It also reduces context length by 35\%, training time by 5.4\%, and inference latency by 30.9\% in wall-clock time. We additionally release the distillation data, prompts, and compressor checkpoint to support reproducibility.
\end{itemize}

\section{Related Works}
\label{asec:related-works}

\paragraph{From RAG to Search Agents.}
Retrieval-Augmented Generation~\citep[RAG,][]{Patrick2020rag} grounds LLM generation in passages retrieved from an external knowledge store~\citep{xu2025surveyofmodelarchitectures,gao2023retrievalaugmentedgeneration}. Classical RAG follows a single retrieve-then-generate pass; subsequent work moves toward multi-step variants in which retrieval is interleaved with reasoning. IRCoT~\citep{trivedi2023ircot} interleaves retrieval with chain-of-thought steps, and Self-RAG~\citep{asai2024self} supervises the generator to decide adaptively when to retrieve.

These methods anticipate, but do not yet realize, the modern \emph{search-agent} formulation, in which an LLM is trained to act as an agent that issues tool calls (here: search queries) and reads back environment observations (retrieved passages) over many turns. The conceptual lineage runs through tool-augmented prompting frameworks such as ReAct~\citep{yao2023react} and Toolformer~\citep{schick2023toolformer}, where reasoning, tool invocation, and observation are interleaved in a single decoding stream. Recent reasoning-aware RL methods~\citep{guo2024deepseek,hurst2024gpt4osystemcard,guo2025deepseek} have made it practical to train such agents end-to-end: ReSearch~\citep{chen2025learningtoreason}, R1-Searcher~\citep{song2025r1searcher}, DeepResearcher~\citep{zheng2025deepresearcher}, and Search-R1~\citep{jin2025searchr1,jin2025searchr1extension} all train an LLM policy under a verifiable reward to interleave search and reasoning across multiple turns. Among these, we focus on Search-R1~\citep{jin2025searchr1} as our reference search-agent backbone; \ours's plug-and-play design generalizes to the others.

\paragraph{Compressing Environment Observations.}
Long, noisy contexts are a generic pathology for LLMs~\citep{jin2025longcontextllmsmeetrag,liu-etal-2024-lost}, but multi-turn search agents make the problem acute: every tool call enlarges the context with raw retrieved passages, and the next reasoning step must read through the accumulated stream. Beyond cost and latency, this stream-growth dynamic interacts poorly with RL training---rollout memory becomes a bottleneck, and the policy is asked to attend over an ever-longer mix of relevant and distracting evidence~\citep{cao2025skyrl,feng2025groupingroup}.

Prior work on context compression spans several families, all developed in the single-pass RAG setting. \emph{Hard-prompt pruning} methods such as LLMLingua~\citep{jiang-etal-2023-llmlingua} and selective-context pruning~\citep{li-etal-2023-compressing} drop low-information tokens to accelerate inference. \emph{Filtering and reranking} approaches train a separate model to discard irrelevant passages---FILCO~\citep{wang-etal-2024-learning-to-filter} learns context filters for RAG, and MAIN-RAG~\citep{chang-etal-2025-main} uses a judge model to filter retrievals. \emph{Soft-prompt compression} methods compress context into latent vectors~\citep{Rae2020Compressive,chevalier-etal-2023-adapting,ge2024incontextautoencoder}. \emph{Robustness-oriented} approaches instead train the generator to tolerate noisy context~\citep{yoran2024makingretrievalaugmented,fang-etal-2024-noiserag}. The most closely related work, RECOMP~\citep{xu2024recomp}, trains a dedicated summarization module for single-turn retrieve-then-read pipelines.

\ours differs from these in \emph{setting} rather than mechanism, and the setting dictates the design. Hard-prompt and soft-prompt compression both sacrifice the abstractive, human-readable form we need to keep the agent's interleaved reasoning trace auditable~\citep{Lanham2023MeasuringFI,baker2025monitoring,korbak2025chain}; filtering and reranking methods retain raw passages and therefore do not shorten any individual observation; RECOMP is single-pass and does not contend with the per-turn observation stream of a search agent, where the compressor must run on every tool call and remain decoupled from the policy gradient. A concurrent work, ReSum~\citep{wu2025resum}, leverages a 32B summarizer for long-horizon web browsing, targeting a different scale and task setting. In contrast, \ours integrates a lightweight, frozen, abstractive observation compressor directly into the RL training loop of a search agent, demonstrating that learned compression under these constraints improves both efficiency and accuracy. We view experimental comparison against the above single-pass methods, adapted to the per-tool-call frozen-during-RL setting, as an important direction for future work.

\section{Background: Search-R1 as a Search Agent}
\label{asec:background-searchr1}
Search-R1~\citep{jin2025searchr1} formulates a search agent as a reinforcement learning problem in which the policy alternates between generating reasoning spans and issuing tool calls to an external retriever. For an input question $x$, a rollout interleaves model-generated reasoning with environment observations returned by the tool and ends with a final answer. Denoting a trajectory by $y=\{s_1,o_1,s_2,o_2,\ldots,s_T,a\}$, where $s_i$ is a reasoning span, $o_i$ is the environment observation (retrieved evidence) at step $i$, and $a$ is the final answer, the multi-turn formulation lets the agent iteratively acquire missing evidence rather than relying only on parametric memory.

The policy is optimized with PPO using the final-answer exact-match signal as the task reward together with a KL penalty that keeps the learned policy close to a reference model. In simplified form, the objective can be written as
\[
\max_{\pi_\theta}\; \mathbb{E}_{x\sim\mathcal{D},\,y\sim\pi_\theta(\cdot\mid x;\mathcal{R})}\big[r_\phi(x,y)\big]
\;-\; \beta\, D_{\mathrm{KL}}\!\big(\pi_\theta(y\mid x;\mathcal{R})\,\|\,\pi_{\mathrm{ref}}(y\mid x;\mathcal{R})\big),
\]
where $r_\phi(x,y)$ is the task reward and $\pi_{\mathrm{ref}}$ is the reference policy. For brevity, we omit the full clipped surrogate PPO objective here and present only the high-level RL formulation; the detailed PPO form is standard and not central to our contribution. Importantly, only tokens generated by the agent---reasoning spans, tool-call queries, and the final answer---are optimized, while observation tokens are masked out from the loss. This setup lets Search-R1 jointly learn to search and reason.

A key limitation, however, is that Search-R1 appends each environment observation $o_i$ to the agent's context verbatim after every tool call. The raw observations are typically verbose, redundant, or only partially relevant to the query; concatenating them across turns lengthens the context and dilutes the signal available for downstream reasoning. This is the limitation \ours targets: rather than feeding raw observations back to the agent, we route each one through a learned compressor that condenses it into concise, query-focused evidence before re-injection.

\section{Proposed Method}
\label{sec:proposed-method}

We present \ours, which addresses the limitation identified above---verbose, noisy environment observations accumulating in the agent's context---by interposing a learned observation compressor between the search tool and the agent. The compressor is trained in two stages and remains frozen during RL training of the agent.

Our framework comprises (1) a summarizer that filters and condenses each observation returned by the search tool, and (2) an integration strategy that routes every tool call through the summarizer before its output is appended to the agent's context (see~\cref{fig:pipeline}). This design enables multi-turn search-and-reason rollouts whose context length stays bounded turn over turn.

\subsection{Summarizer Training}
\label{subsec:summarizer-training}
The summarizer is trained in two stages. The first stage equips it with relevance detection over retrieved passages; the second aligns it with high-quality summarization preferences distilled from a stronger teacher model.

\paragraph{Stage 1: Relevance pretraining on MS MARCO.}
Our goal is to enable the summarizer to identify relevant information within a noisy observation. In the MS~MARCO question answering dataset~\citep{bajaj2016msmarco}, at most one passage per query is labeled as relevant.
Given a query $x_i$ and ten candidate passages $D_i=\{d_{i1},\ldots,d_{i10}\}$, the relevance-pretraining task is to correctly classify the ground-truth relevant passage. We initialize with a lightweight model $M_s$~(Qwen2.5-3B-Instruct) augmented with a linear classification head, scoring each $d_{ij}$ conditioned on $x_i$ under a cross-entropy loss. This equips the summarizer with a strong relevance prior for later generative training. As we show in~\cref{subsec:result_analysis}, this step is critical for the final search-agent performance.

\paragraph{Stage 2: Summarization training via distillation.}
Having equipped the summarizer model with the basic ability to identify relevant information, we now aim to improve its summarization capability.
Recall our principle is to design a modular, plug-and-play summarizer that can meet different summary desiderata.
We draw inspiration from prior works on abstract summarization and its evaluation~\citep{kryscinski2020evaluatingthefactualconsistency,fabbri2021summeval,zhu2021enhancingfactualconsistency,luo2023chatgptfactualinconsistencyevaluator} and propose a paradigm to train the summarizer to summarize the retrieved documents for different aspects. 
We refer to our approach as \textit{multi-aspect summaries}. We design six guiding aspects: \emph{factual correctness}, \emph{completeness}, \emph{coverage}, \emph{coherence}, \emph{clarity}, and \emph{logicality} following the literature, then craft corresponding training data for optimization.

Specifically, we use the seed queries from NQ and HotpotQA datasets, consistent with \ours’s training distribution to avoid leakage. For each seed question $x_i$, we collect the query sets $X_i$ and associated document sets $\mathcal{D}_i$ from Search-R1's rollouts. We prompt a proprietary model GPT-4o-mini for a multi-document query-focused summary~\citep{xu2020coarse} under the aforementioned aspects (prompt templates in~\cref{asec:prompt-templates}). Notably, we design our prompt to include a general instruction and the specific aspect, instructing the model to generate a high-quality summary while optimizing for one specific aspect. 
Finally, we obtain 468k HotpotQA and 1.0M NQ summaries, forming the distillation mixture (more details in~\cref{asec:distillation-data}).

The summarizer, initialized from Stage 1 with the classification head discarded, is trained to reproduce the teacher outputs via teacher forcing~\citep{kim-rush-2016-sequence}. We train on the mixture of all six aspects jointly.
This distillation stage transfers multi-aspect supervision into a small, efficient compressor that drops in as a frozen module in the RL-trained search agent.

\subsection{Integrating the \ours Framework}
\label{sec:integration}

\input{tables/dataset}

We implement \ours by augmenting the Search-R1 search-agent pipeline with a summarization step. In stock Search-R1, retrieved passages are concatenated into the agent's context verbatim after each tool call. In \ours, every retrieved observation is instead routed through the frozen summarizer, which condenses it into a concise yet human-readable summary before it is appended to the context (\cref{fig:pipeline}). For experiments, we exclusively use the \emph{clarity} setting of the summarizer. Among the six aspects, Clarity produces the shortest summaries (581 characters on average, compared to 593--1006 for the other aspects; see~\cref{asec:distillation-data}), while encouraging concise, unambiguous language that preserves key facts---properties well-suited to serving as the per-tool-call observation handed to the agent (see case studies in~\cref{asec:case-study}).

We highlight \ours's flexibility: because the summarizer is trained under a multi-aspect setup, an operator can swap in a different aspect (e.g., \textit{factuality}, \textit{completeness}) at inference time without retraining---making \ours adaptable to other search-agent tasks where the desired observation form differs.

\input{tables/main_em}

\paragraph{Training.}
We train the agent on the mixture of NQ and HotpotQA training sets using the same hyperparameters as Search-R1, with two modifications enabled by per-tool-call compression: (1) the number of retrieved passages per tool call is increased from 3 to 5, and (2) the maximum number of tool calls per rollout is extended from 3 to 5. Because the summarization step keeps observations short, the agent can afford a deeper search-and-reason horizon at the same context budget. As in the baseline, we optimize the policy with Proximal Policy Optimization~\citep[PPO,][]{schulman2017proximal}. The full algorithm is given in~\cref{alg:searchr1-sum}, with additional details in~\cref{asec:algorithmic}.

\paragraph{Outcome.}
By compressing each environment observation before re-injection, \ours produces shorter agent contexts, lets the agent reach an answer in fewer tool calls, and achieves lower inference latency. As we observe in case studies (\cref{asec:case-study}), the summarizer also reduces distraction from irrelevant content, improving accuracy especially on multi-hop questions (\cref{subsec:result_analysis}).

\section{Experiments}
\label{sec:experiments}

\subsection{Experimental Setup}
\label{subsec:experimental_setup}

Our setup follows the Search-R1 pipeline for a fair comparison.

\paragraph{Baselines.}
\label{asec:baselines}
We primarily compare against Search-R1, a canonical RL-trained search agent. Search-R1 trains a search-and-reason interleaved LLM agent with reinforcement learning using both PPO and its variant GRPO~\citep{schulman2017proximal,shao2024deepseekmath}. All agents are trained on a mixture of Natural Questions (NQ)~\citep{kwiatkowski2019natural} and HotpotQA~\citep{yang2018hotpotqa}, using the Wikipedia-18 dump as the knowledge corpus, with a unified preprocessing pipeline. For completeness, we also report representative prompting-based baselines (Direct Inference and Chain-of-Thought prompting~\citep{wei2022chain}), retrieval-based baselines (RAG~\citep{Patrick2020rag}, IRCoT~\citep{trivedi2023ircot}, Search-o1~\citep{li2025searcho1}), and training-based baselines (SFT, RL without a search tool~\citep{guo2025deepseek}, rejection sampling without a search tool~\citep{ahn2024rejectsample}). All baseline results are from~\citet{jin2025searchr1}.

\input{tables/appendix_base_vs_instruct}

\paragraph{Backbones and datasets.}
We use Qwen2.5-\{3B, 7B\}-Base~\citep{yang2024qwen25} as the backbone for the search agent's policy. In line with prior works, Base models paired with PPO yield the strongest performance among available variants. Evaluation spans seven QA benchmarks: NQ~\citep{kwiatkowski2019natural}, TriviaQA~\citep{joshi2017triviaqa}, PopQA~\citep{mallen-etal-2023-trust}, HotpotQA~\citep{yang2018hotpotqa}, 2Wiki~\citep{ho-etal-2020-constructing}, Musique~\citep{trivedi2022musique}, and Bamboogle~\citep{press2023bamboogle}. These cover open-domain QA and multi-hop QA, jointly assessing factual accuracy and the depth of multi-tool-call reasoning. For relevance pretraining, we use the MS MARCO QA train split released by~\citet{jin2024FlashRAG}.

\paragraph{Distillation data construction.}
\label{asec:distillation-data}
We build summarizer training data from Search-R1 rollouts using the official checkpoint\footnote{\url{PeterJinGo/SearchR1-nq\_hotpotqa\_train-qwen2.5-7b-em-ppo}}. For each query in NQ and HotpotQA, we collect intermediate search queries, deduplicate them, retrieve top-5 passages with E5-base-v2~\citep{wang2022text}, and form query-document pairs. We then prompt GPT-4o-mini to generate query-focused multi-document summaries for six aspects, yielding 468{,}547 triplets from HotpotQA and 1{,}002{,}329 from NQ. The average summary lengths (in characters) are 581 for Clarity, 593 for Factual Correctness, 714 for Completeness, 774 for Logicality, 875 for Coherence, and 1006 for Coverage; we therefore use the \emph{Clarity} setting in all experiments.

\paragraph{Evaluation.}
We report exact match (EM) as the primary metric. To assess efficiency, we additionally track average context length, inference time, and number of reasoning turns, which allows us to disentangle gains from accuracy versus efficiency.

\paragraph{Implementation and hyperparameters.}
\label{asec:hyperparameters}
Our system uses PyTorch, veRL-v0.1~\citep{sheng2024hybridflow}, vLLM~\citep{kwon2023efficient}, Huggingface Transformers, and LlamaFactory~\citep{zheng-etal-2024-llamafactory}. We train the summarizer with LlamaFactory for both relevance pretraining and distillation using an effective batch size of 16, learning rate $1\times10^{-4}$, 5\% warmup, 1 epoch, bf16, LoRA (rank 8, $\alpha=32$), and maximum sequence length 2048. During RL training, the summarizer is frozen. We train \ours with PPO using a global batch size of 512, actor/critic learning rates of $10^{-6}$/$10^{-5}$, clip ratio 0.2, max prompt/response lengths of 4096/500 tokens, and bfloat16 training. The retriever is frozen E5-base-v2 with FAISS acceleration, and we retrieve top-5 passages per turn. Training runs on 4 H100 GPUs for the 3B model and 8 H100 GPUs for the 7B model; rollout sampling uses top-$p=0.9$, top-$k=40$, and temperature $0.7$.

\paragraph{Memory footprint.} The frozen Qwen2.5-3B summarizer is co-located with the policy on the same GPUs in bf16 (approximately 6\,GB), and is invoked once per retrieval step on the concatenation of top-$5$ retrieved passages. Summarizer inference is overlapped with the search call in our pipeline, so the additional GPU-memory and latency overhead is small relative to the policy itself; the net effect---reduced policy context length---dominates and yields the wall-clock improvements reported in~\cref{fig:efficiency-inference}.

\paragraph{Discussion of the retrieval/turn-budget configuration.} Following Search-R1's setup, the baseline uses top-$3$ retrieved passages and up to $3$ turns. We train \ours with top-$5$ passages and up to $5$ turns, made affordable by per-step compression. To isolate the contribution of compression itself from the increased budget, our relevance-pretraining ablation (\cref{tab:main_em}) holds the budget fixed and shows that removing Stage~1 collapses 7B training even at the same top-$5$ / $5$-turn setting---indicating that a deeper budget alone is not sufficient and that the two-stage summarizer is doing meaningful work.

\subsection{Results and Analysis}
\label{subsec:result_analysis}

\input{figures/inference_efficiency}
\paragraph{Performance results.}
\cref{tab:main_em} shows that \ours consistently outperforms Search-R1 across backbones and datasets. With Qwen2.5-7B-Base, \ours raises average EM from 0.431 to 0.444. Improvements are even more pronounced on the 3B agent, where EM increases from 0.303 to 0.347 (a 14.5\% relative gain), underscoring the value of observation compression for compact agents.

The largest gains appear on multi-hop QA benchmarks (HotpotQA, 2Wiki, Musique, and Bamboogle), where the agent must reconcile evidence across multiple tool calls; compressing each observation yields a cleaner running context that better supports multi-step reasoning. \ours also improves performance on single-hop QA (e.g., TriviaQA: 0.675 vs. 0.638), indicating that the benefits extend beyond multi-hop tasks.

\paragraph{Efficiency results.}

\cref{fig:efficiency-inference} compares inference efficiency, normalized to the Search-R1 baseline (1.0x). \ours reduces per-observation length and filters irrelevant content, which in turn lowers the number of tool calls the agent issues to reach an answer. These factors together significantly reduce inference latency in wall-clock time.
\ours also improves training time. Using Qwen2.5-3B-Base + PPO on 4$\times$ H200 GPUs, we measure a 13.9-hour training time for \ours over 500 steps, compared to Search-R1's 14.7 hours---a 5.4\% speedup. Overall, our results show that \ours improves both training speed and inference latency relative to Search-R1, even after accounting for the extra summarization step at every tool call. See~\cref{asec:efficiency-results} for complete efficiency results.

\paragraph{Relevance pretraining.} 
The relevance-pretraining ablation is included in~\cref{tab:main_em}. Replacing the first-stage relevance pretraining with Stage 2 distillation-only training reduces the 3B model average EM from 0.347 to 0.295, below the Search-R1 baseline, and causes the 7B variant to deteriorate sharply (0.304 average EM; results reported at step 300 because training collapses afterward). These results suggest that Stage 1 + Stage 2 training is necessary for stable and effective summarizer learning.

\paragraph{Backbone variants.}
We further evaluate \ours with an Instruct-tuned agent backbone. As shown in~\cref{tab:base-vs-instruct-appendix}, with Qwen2.5-3B-Instruct + PPO, \ours achieves an average EM of 0.336 compared to 0.325 for Search-R1. Although the margin is smaller than with the Base backbone, the consistent direction of improvement confirms that \ours generalizes across backbone families.

\paragraph{Case studies and faithfulness analysis.}
To contextualize the behavior of \ours's summarizer, an author manually examined 100 NQ samples (\cref{asec:case-study}). 86/100 summaries preserve the information needed to answer the query, while 14/100 exhibit information loss that could in principle harm answering. The lossy cases cluster into four recurring patterns---dropped named entities, missing temporal information, missing numerical values, and altered compound phrases---which we discuss qualitatively in~\cref{asec:case-study}. The 14\% information-loss rate is partially mitigated by the agent's ability to issue follow-up tool calls in subsequent turns; net of this mitigation, \ours yields an overall EM gain across all seven benchmarks. We view a quantitative, dataset-scale evaluation of summarizer faithfulness (using LLM-as-judge or factuality-oriented metrics) as an important next step (\cref{sec:limitations}).

\section{Limitations and Discussion}
\label{sec:limitations}

We discuss limitations of \ours to scope its claims and to motivate directions for future work.

\paragraph{Evaluation metric.} We report exact match (EM) as the primary metric, following Search-R1~\citep{jin2025searchr1} and the standard practice on these seven benchmarks. EM is a coarse signal: it does not separately capture answer faithfulness, evidence recall, or hallucination rate, all of which are particularly relevant for an abstractive compressor. As a partial mitigation, we conduct a manual examination of 100 NQ samples (\cref{asec:case-study}) to characterize information-loss patterns in the summarizer. A more thorough evaluation with LLM-as-judge or factuality-oriented metrics (e.g., FactScore-style decomposition) is left to future work.

\paragraph{Baseline scope.} Our experiments compare against Search-R1~\citep{jin2025searchr1} and the prompting/retrieval\\/training-based baselines reported therein, and ablate the relevance-pretraining stage of \ours itself. We do not experimentally compare against single-pass compression methods such as RECOMP~\citep{xu2024recomp}, LLMLingua~\citep{jiang-etal-2023-llmlingua}, or FILCO~\citep{wang-etal-2024-learning-to-filter}, because adapting these methods to the per-tool-call, frozen-during-RL setting is itself a non-trivial research question---each was designed for single-pass retrieve-then-read pipelines and would require re-training and re-integration to operate on every environment observation under PPO. We position \ours as the first instantiation of these design constraints in an RL-trained search agent (\cref{sec:intro}); a head-to-head study against compression-method variants under a unified multi-turn agent protocol is an important direction we leave to future work.

\paragraph{Scaling and gain magnitude.} Gains over Search-R1 are larger on the 3B agent (14.5\% relative average EM) than on the 7B agent (3.0\% relative). One plausible explanation is that larger policies tolerate noisier observations better, so the marginal benefit of compression diminishes; another is that the summarizer (Qwen2.5-3B) becomes capacity-limited relative to a 7B+ policy. Whether \ours continues to help 13B/70B-scale agents, and whether scaling the summarizer alongside the agent recovers the gain, are open questions we cannot answer within our present compute budget.

\paragraph{Information loss in abstractive summaries.} Among 100 manually examined NQ samples (\cref{asec:case-study}), 14 exhibit information loss in the summary that could in principle harm answering. We characterize four recurring failure patterns: dropped named entities, missing temporal/numeric details, and altered compound phrases. The agent's ability to issue follow-up tool calls in subsequent turns partially compensates for these losses, but the trade-off between conciseness and fidelity is fundamental to abstractive compression and warrants targeted work (e.g., entity-preserving training objectives, copy mechanisms).

\paragraph{Distillation source.} The summarizer is distilled from GPT-4o-mini outputs, which inherits any biases or factual errors of that teacher. We do not currently audit the distilled corpus for residual hallucination; a teacher-ensembling or self-consistency filter would be a natural follow-up.

\section{Conclusion and Future Works}

We presented \ours, a framework that addresses the efficiency and performance challenges of RL-trained search agents by compressing environment observations on the fly. By inserting a dedicated summarization module---trained via relevance pretraining followed by multi-aspect distillation, and frozen during RL---between the search tool and the agent, \ours converts verbose tool outputs into concise, factual observations that better align with the agent's downstream reasoning. Integrated into the Search-R1 search-agent pipeline, \ours yields consistent improvements across seven QA benchmarks: it shortens the running context, accelerates both training and inference, and lifts reasoning accuracy, particularly on multi-hop tasks that demand reconciling evidence across many tool calls.
As search agents are pushed toward longer-horizon, multi-tool settings---web browsing, scientific discovery, computer use---observation compression becomes a practical necessity rather than a useful optimization. \ours provides a concrete, modular approach to this problem, and we see extending it to richer tool ecosystems and more diverse environments as a promising direction for future work.

\bibliographystyle{plainnat}
\bibliography{references}

\appendix
\section{Algorithmic Details}
\label{asec:algorithmic}
We show the detailed algorithm in~\cref{alg:searchr1-sum}. \textcolor{OrangeRed}{Colored text} denotes modifications from the original Search R1 algorithm. 
\begin{algorithm}[t]
\caption{Search-R1 rollout with summarization-augmented retrieval}
\label{alg:searchr1-sum}
\begin{algorithmic}[1]
\Require Query $x$; policy $\pi_\theta$; retriever $\mathcal{R}$; summarizer $S$; action budget $B$
\Ensure Final response $y$
\State Initialize rollout $y \gets \varnothing$ and action count $b \gets 0$
\While{$b < B$}
  \State Initialize current action segment $y_b \gets \varnothing$
  \Loop
    \State Sample token $y_t \sim \pi_\theta(\cdot \mid x, y + y_b)$
    \State Append token: $y_b \gets y_b + y_t$
    \If{$y_t \in \{\texttt{</search>}, \texttt{</answer>}, \texttt{<eos>}\}$}
      \State \textbf{break}
    \EndIf
  \EndLoop
  \State Append segment to rollout: $y \gets y + y_b$
  \If{$y_b$ contains \texttt{<search>} $\cdots$ \texttt{</search>}}
    \State Parse search query:
    \Statex\hspace{\algorithmicindent}$q \gets \mathrm{Parse}(y_b, \texttt{<search>}, \texttt{</search>})$
    \State Retrieve documents $d \gets \mathcal{R}(q)$
    \State \textcolor{OrangeRed}{Summarize retrieved documents $d' \gets S(d)$}
    \State Inject evidence into the rollout:
    \Statex\hspace{\algorithmicindent}$y \gets y + \texttt{<information>}\,\textcolor{OrangeRed}{d'}\,\texttt{</information>}$
  \ElsIf{$y_b$ contains \texttt{<answer>} $\cdots$ \texttt{</answer>}}
    \State \Return $y$
  \Else
    \State Append fallback message to the rollout:
    \Statex\hspace{\algorithmicindent}$y \gets y +$ ``My action is not correct. Let me rethink.''
  \EndIf
  \State Increment action count: $b \gets b + 1$
\EndWhile
\State \Return $y$
\end{algorithmic}
\end{algorithm}

For completeness, we present the PPO objective used for policy optimization with the summarizer-augmented retriever $\mathcal{R}_{\text{sum}}$.
\begin{equation}
\begin{aligned}
\mathcal{J}_{\text{PPO}}(\theta)
=\;& \mathbb{E}_{x\sim\mathcal{D},\, y\sim\pi_\theta(\cdot\mid x;\mathcal{R}_{\text{sum}})} \Bigg[
\frac{1}{|y|} \sum_{t=1}^{|y|} \mathbb{I}(y_t)\cdot \min\!\Big( r_t(\theta) A_t,\,
\operatorname{clip}\big(r_t(\theta), 1\!-
\epsilon, 1\!+\epsilon\big) A_t \Big)
\Bigg]
\end{aligned}
\label{eq:ppo}
\end{equation}
\begin{equation}
\label{eq:rt}
r_t(\theta) =
\frac{\pi_\theta\big(y_t \mid x, y_{<t}; \mathcal{R}_{\text{sum}}\big)}
{\pi_{\text{old}}\big(y_t \mid x, y_{<t}; \mathcal{R}_{\text{sum}}\big)}.
\end{equation}
Here, $\pi_\theta$ and $\pi_{\text{old}}$ denote the current and previous policy models; $\mathbb{I}(y_t)$ equals $1$ for LLM-generated tokens and $0$ for retrieved tokens; $A_t$ is the advantage estimate; and $\epsilon$ is the clipping parameter.

\section{Detailed Efficiency Results}
\label{asec:efficiency-results}
We report detailed inference-time efficiency results in~\cref{tab:efficiency-appendix}.
\input{tables/appendix_efficiency}

\section{Prompt Templates}
\label{asec:prompt-templates}
The six aspect and their aspect explanation with the prompt template is shown below.

\begin{itemize}
  \item \textbf{Clarity}
  Write in a clear, accessible manner that is easy to understand. Use simple, direct language and avoid jargon or overly complex sentences. Present information in a straightforward way that makes the key points immediately apparent to the reader. Ensure that each statement is unambiguous and easy to follow.
  \item \textbf{Coherence} 
  Create a logically coherent and well-structured summary that flows naturally from one point to the next. Use clear transitions, logical connections, and a consistent narrative structure. Ensure that ideas are presented in a logical sequence that makes sense to the reader, with each piece of information building upon the previous one.
  \item \textbf{Completeness} 
  Make sure that the support context includes all major facts from the retrieved documents that are needed to answer the user's question. Do not omit important information, even if it seems implicit. The summary should be as thorough as possible within a compact form.
  \item \textbf{Coverage}
  Ensure comprehensive coverage of all relevant information from the retrieved documents. Include all key facts, data points, examples, and supporting details that could be useful for answering the question. Avoid omitting important information even if it seems redundant, as comprehensive coverage is prioritized over brevity.
  \item \textbf{Factual Correctness}
  Ensure that every statement in the support context is factually accurate and directly supported by the retrieved documents. Do not include any information that is inferred, assumed, or fabricated. Avoid hallucinations, exaggerations, or unsupported claims.
  \item \textbf{Logicality} 
  Present information in a logically sound manner with clear reasoning and valid conclusions. Ensure that cause-and-effect relationships are properly established, that arguments are well-structured, and that conclusions follow logically from the presented evidence. Avoid logical fallacies and ensure that the information flows in a way that makes logical sense.
\end{itemize}

\input{figures/prompt_template}

Our selection of aspects was motivated by an array of prior works on abstract summarization. Mainly,~\citet{fabbri2021summeval} design 4 evaluation dimensions of abstract summarization to be used in human annotation, these annotations are subsequently used for meta evaluation of evaluation metrics.
\begin{itemize}
    \item \textbf{Coherence}: the collective quality of all sentences. ``The summary should be well structured and well organized, instead of just being a heap of related information''.
    \item \textbf{Consistency}: the factual alignment between the summary and the summarized source.
    \item \textbf{Fluency}: the quality of individual sentences. ``Sentences should have no formatting problems, capitalization errors or obviously ungrammatical sentences that make the text difficult to read.''
    \item \textbf{Relevance}: selection of important context from the source. The summary should include only important information from the source document.
\end{itemize}
Our six aspects are connected to these four dimensions. 1/ Clarity reflects \citet{fabbri2021summeval}’s Relevance. 2/ Coherence is directly mapped to \citet{fabbri2021summeval}’s Coherence. 3/ Completeness optimizes for the Recall of important information, connecting to \citet{fabbri2021summeval}’s Relevance. 4/ Coverage is similar to Completeness, but focuses on the “query-focused” aspect of RAG. 5/ Factual Correctness is a direct reflection of \citet{fabbri2021summeval}’s Consistency. 6/ Logicality is similar to Coherence, but focuses more on the inherent logical chain of the documents, optimized for RAG queries that requires reasoning.

\section{Case Study and Summarizer Failure Patterns}
\label{asec:case-study}
\input{figures/case_study}
We additionally show a case study. The summarizer cut a large, messy set of documents down to a short, clear answer. It reduced the text from 1,821 characters to 339 and removed most of the irrelevant material, including an entirely different TV show. As a result, the useful information went from about a third of the content to almost all of it. The final summary puts the episode count right up front and drops distracting details like DVD extras and social-media reactions. Overall, it makes the information much easier to read and understand to benefit the search agent.

An author manually went through 100 samples from the NQ dataset~\citep{kwiatkowski2019natural}. Among 100 samples, we identified 14 samples where the summarizer leads to a certain degree of information loss, despite the presence of relevant information in the retrieved document, while 86 summaries retain the necessary information for the RAG generator (the policy) to answer the query. The breakdown is summarized in~\cref{tab:faithfulness}.

\begin{table}[h]
\centering
\small
\caption{Manual faithfulness audit on 100 NQ samples. ``Information preserved'' indicates the summary retains the evidence needed to answer the query; ``information loss'' indicates relevant content is omitted from the summary despite being present in the retrieved document.}
\label{tab:faithfulness}
\begin{tabular}{lc}
\toprule
Outcome & \# Samples (out of 100) \\
\midrule
Information preserved & 86 \\
Information loss      & 14 \\
\midrule
\quad of which: dropped named entity        & (qualitative; see below) \\
\quad of which: missing temporal information & (qualitative; see below) \\
\quad of which: missing numerical value      & (qualitative; see below) \\
\quad of which: altered compound phrase      & (qualitative; see below) \\
\bottomrule
\end{tabular}
\end{table}

Among the 14 lossy samples, we identified the following representative error patterns:
\begin{itemize}
    \item \textbf{Named entity}: the query asks about a specific named entity, e.g., Yaya Tour\'e, Charles Darwin. These mentions of these named entities are lost in the summarized evidence.
    \item \textbf{Temporal information}: the query asks about time of a specific event. The temporal information is lost in the summarization process.
    \item \textbf{Numerical information}: the query asks about a precise numerical value about measurements or quantities, and such information is omitted by the summarizer.
    \item \textbf{Compound definitions}: the groundtruth is a specific compound phrase, e.g., ``hit points or health points'', but the summarizer leaves ``or'' out.
\end{itemize}

We note that the summarization errors often do not translate to the failure of the search agent, as the search agent is trained to conduct multiple rounds of search and reasoning\,---\,specifically to rephrase the query and search more when no information in the retrieved documents supports answering the question. Therefore, \ours is still effective for improving the RAG performance, as shown in~\cref{subsec:result_analysis}.

\end{document}

%% file: figures/pipeline.tex
\begin{figure*}[t]
    \centering
    \includegraphics[width=\textwidth]{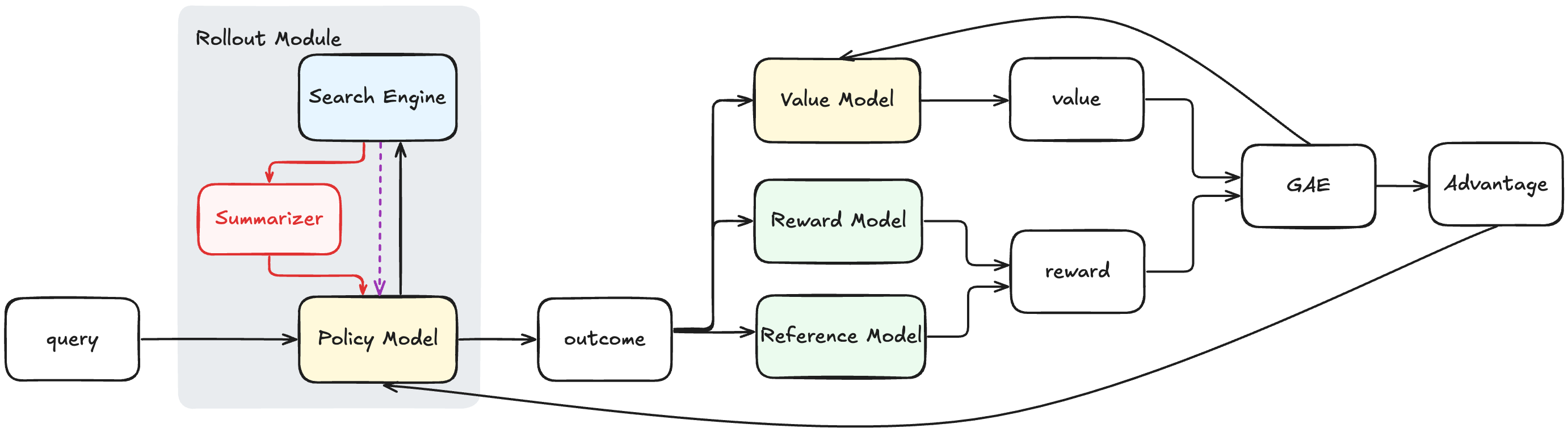}
    \caption{Training pipeline of our method. We reuse the color scheme from Search-R1. \textcolor{Yellow}{Yellow} denotes models to be updated in training. \textcolor{ForestGreen}{Green} denotes model kept frozen in training. \textcolor{blue}{Blue} denotes the environment. In the \textcolor{gray}{rollout module}, instead of directly using retrieval results from the search engine (\textcolor{Plum}{dashed line}), an additional \textcolor{OrangeRed}{summarization model} is used to condense the retrieved information and remove noises from document sources. This way we reduce the context length and achieve efficient and effective rollout in both training and inference.}
    \label{fig:pipeline}
\end{figure*}

%% file: tables/dataset.tex
\begin{table}[t]
\centering
\caption{QA Dataset statistics. $\dag$ denotes in-domain trainset and $^*$ denotes OOD dataset.}
\vspace{-10pt}
\label{tab:dataset-stats}
\resizebox{0.8\textwidth}{!}{
\begin{tabular}{lllll}
\toprule
Dataset & \# Train & \# Val & \# Test & Task \\
\midrule
Natural Questions$^\dag$~\citep{kwiatkowski2019natural} & 79{,}168 & 8{,}757 & 3{,}610 & QA \\
TriviaQA$^*$~\citep{joshi2017triviaqa} & 78{,}785 & 8{,}837 & 11{,}313 & QA \\
PopQA$^*$~\citep{mallen-etal-2023-trust} & --       & --      & 14{,}267 & QA \\
HotpotQA$^\dag$~\citep{yang2018hotpotqa} & 90{,}447 & 7{,}405 & --       & Multi-hop QA \\
2WikiMultihopQA$^*$~\citep{ho-etal-2020-constructing} & 15{,}920 & 1{,}986 & 1{,}996  & Multi-hop QA \\
MuSiQue$^*$~\citep{trivedi2022musique} & 19{,}938 & 2{,}417 & --       & Multi-hop QA \\
Bamboogle$^*$~\citep{press2023bamboogle} & --       & --      & 125      & Adversarial QA \\
\bottomrule
\end{tabular}
}
\end{table}

%% file: tables/main_em.tex
\begin{table*}[t]
\centering
\caption{Performance comparison of different methods on selected QA datasets, including the relevance-pretraining ablation. $^\dagger$ denotes in-domain trainset and $^*$ denotes OOD dataset. For RECON, we report both Stage 2 Distillation-only and RECON (Stage 1 + Stage 2). For the 7B model, the Stage 2 Distillation-only configuration is reported at step 300 because training collapses afterward. Best overall numbers on each dataset are \textbf{bold}, best in-category numbers are \underline{underlined}.}
\label{tab:main_performance_em}
\resizebox{\textwidth}{!}{% To make the table fit the page width if it's too wide
\begin{tabular}{@{}lcccccccc@{}}
\toprule
\multirow{2}{*}{\textbf{Methods}} & \multicolumn{3}{c}{\textbf{General QA}} & \multicolumn{4}{c}{\textbf{Multi-Hop QA}} & \multirow{2}{*}{\textbf{Avg.}} \\
\cmidrule(r){2-4} \cmidrule(lr){5-8}
& NQ$^\dagger$ & TriviaQA$^*$ & PopQA$^*$ & HotpotQA$^\dagger$ & 2wiki$^*$ & MuSiQue$^*$ & Bamboogle$^*$ & \\
\midrule
\multicolumn{9}{l}{\textit{Baseline methods}} \\
Direct Inference & 0.134 & 0.408 & 0.140 & 0.183 & 0.250 & 0.031 & 0.120 & 0.181 \\
CoT & 0.048 & 0.185 & 0.054 & 0.092 & 0.111 & 0.022 & 0.232 & 0.106 \\
IRCoT & 0.224 & 0.478 & 0.301 & 0.133 & 0.149 & 0.072 & 0.224 & 0.239 \\
Search-o1 & 0.151 & 0.443 & 0.131 & 0.187 & 0.176 & 0.058 & 0.296 & 0.206 \\
RAG & 0.349 & 0.585 & 0.392 & 0.299 & 0.235 & 0.058 & 0.208 & 0.304 \\
SFT & 0.318 & 0.354 & 0.121 & 0.217 & 0.259 & 0.066 & 0.112 & 0.207 \\
R1-base & 0.297 & 0.539 & 0.202 & 0.242 & 0.273 & 0.083 & 0.296 & 0.276 \\
R1-instruct & 0.270 & 0.537 & 0.199 & 0.237 & 0.292 & 0.072 & 0.293 & 0.271 \\
Rejection Sampling & \underline{0.360} & \underline{0.592} & \underline{0.380} & \underline{0.331} & \underline{0.296} & \underline{0.123} & \underline{0.355} & \underline{0.348} \\
\midrule
\multicolumn{9}{l}{\textit{Qwen2.5-3B-Base + PPO}} \\
Search-R1 & 0.406 & 0.587 & \underline{0.435} & 0.284 & 0.273 & 0.049 & 0.088 & 0.303 \\
\ours (Stage 2 Distillation-only) & 0.400 & 0.575 & 0.409 & 0.271 & 0.230 & 0.057 & 0.121 & 0.295 \\
\ours (Stage 1 + Stage 2) & \underline{0.440} & \underline{0.612} & 0.425 & \underline{0.326} & \underline{0.319} & \underline{0.082} & \underline{0.226} & \underline{0.347} \\
\midrule
\multicolumn{9}{l}{\textit{Qwen2.5-7B-Base + PPO}} \\
Search-R1 & 0.480 & 0.638 & \underline{\textbf{0.457}} & 0.433 & 0.382 & 0.196 & 0.432 & 0.431 \\
\ours (Stage 2 Distillation-only) & 0.296 & 0.556 & 0.421 & 0.305 & 0.286 & 0.128 & 0.127 & 0.304 \\
\ours (Stage 1 + Stage 2) & \underline{\textbf{0.493}} & \underline{\textbf{0.675}} & 0.454 & \underline{\textbf{0.445}} & \underline{\textbf{0.392}} & \textbf{0.206} & \underline{\textbf{0.446}} & \underline{\textbf{0.444}} \\

\bottomrule
\end{tabular}
}
\label{tab:main_em}

\end{table*}

%% file: tables/appendix_base_vs_instruct.tex
\begin{table*}[t]
\centering
\caption{Ablation studies of Base vs Instruct using 3B scale models.  Best overall numbers on each dataset are \textbf{bold}, best in-category numbers are \underline{underlined}.}
\vspace{-10pt}
\label{tab:base-vs-instruct-appendix}
\resizebox{0.8\textwidth}{!}{
\begin{tabular}{@{}lcccccccc@{}}
\toprule
Method & NQ & TriviaQA & PopQA & HotpotQA & 2Wiki & Musique & Bamboogle & Avg. \\
\midrule
\multicolumn{9}{l}{\textit{Qwen2.5-3B-Base + PPO}} \\
Search-R1 & 0.406 & 0.587 & \underline{\textbf{0.435}} & 0.284 & 0.273 & 0.049 & 0.088 & 0.303 \\
\ours & \underline{\textbf{0.440}} & \underline{\textbf{0.612}} & 0.425 & \underline{0.326} & \underline{\textbf{0.319}} & \underline{0.082} & 0.226 & \underline{\textbf{0.347}} \\
\midrule
\multicolumn{9}{l}{\textit{Qwen2.5-3B-Instruct + PPO}} \\
Search-R1 & 0.341& 0.545& 0.378& 0.324& \underline{\textbf{0.319}} & 0.103& 0.264& 0.325\\
\ours & \underline{0.353} & \underline{0.551} & \underline{0.381} & \underline{\textbf{0.359}} & 0.292 & \textbf{0.109} & \underline{\textbf{0.304}} & \underline{0.336} \\
\bottomrule
\end{tabular}
}
\end{table*}

%% file: figures/inference_efficiency.tex
\begin{figure}[t]
    \centering
    \includegraphics[width=0.6\textwidth]{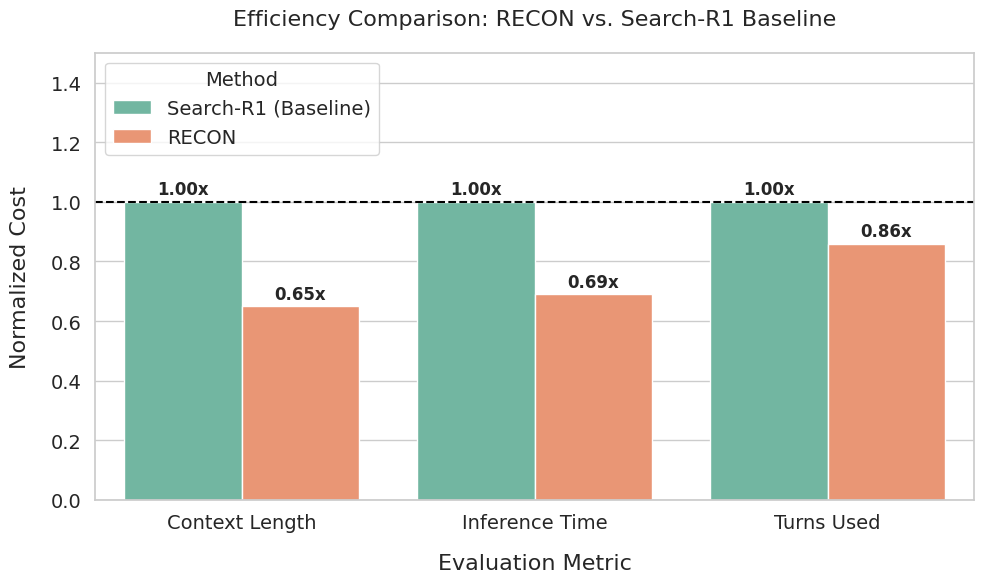}
    \caption{Inference efficiency of \ours vs. Search-R1, with \textit{Qwen2.5-7B-Base + PPO}, normalized to the Search-R1 baseline (1.0$\times$). We report average context length ($\downarrow$), inference wallclock time per query ($\downarrow$) and number of search turns ($\downarrow$) over 7 datasets. Absolute per-dataset values are reported in~\cref{tab:efficiency-appendix}.}
    \label{fig:efficiency-inference}
\end{figure}

%% file: tables/appendix_efficiency.tex
\begin{table*}[t]
\centering
\caption{Comparison of context length (average sequence length), average inference time (s), and turns used (average search count) for tested Search-R1 and proposed \ours. We evaluate \textit{Qwen2.5-7B-Base + PPO}, with the official checkpoint from~\citep{jin2025searchr1}.}
\label{tab:efficiency-appendix}
\resizebox{0.85\textwidth}{!}{
\begin{tabular}{lcccccccc}
\toprule
Method & NQ & TriviaQA & PopQA & HotpotQA & 2Wiki & Musique & Bamboogle & Avg. \\
\midrule
\multicolumn{9}{l}{\textit{Context Length (average sequence length) $\downarrow$}} \\
\quad Search-R1    & 928.5 & 831.6 & 542.9 & 1087.3 & 1123.5 & 1321.0 & 803.1 & 948.3\\
\quad RECON  & 620.7 & 554.4 & 380.5 & 696.0 & 725.3 & 825.6 & 535.4 & 619.7\\
\midrule
\multicolumn{9}{l}{\textit{Average Inference Time (s) $\downarrow$}} \\
\quad Search-R1    & 31.2 & 21.9 & 13.9 & 30.6 & 29.9 & 46.8 & 27.2 & 28.8\\
\quad RECON  & 22.4 & 15.6 & 10.0 & 21.2 & 19.8 & 31.1 & 19.2 & 19.9\\
\midrule
\multicolumn{9}{l}{\textit{Turns Used (average search count) $\downarrow$}} \\
\quad Search-R1    & 2.10 & 1.82& 1.41 & 2.12 & 2.15& 2.57& 2.72& 2.13\\
\quad RECON  & 1.96 & 1.70 & 1.35 & 1.94 & 1.98 & 2.34 & 1.61 & 1.84\\
\bottomrule
\end{tabular}
}
\end{table*}

%% file: figures/prompt_template.tex
\begin{figure*}[h!] % Use 'figure' for a single-column float
    \centering % Center the minipage within the column
    \begin{minipage}{\textwidth} % <-- IMPORTANT: Use \columnwidth
        \begin{tcblisting}{
          colback=gray!5,
          colframe=blue!70!black,
          listing only,
          left=1mm, right=1mm, top=1mm, bottom=1mm,
          title=Summarization Prompt Template
        }
You are a helpful assistant in a retrieval-augmented question-answering system.

You will be given:
- A user question
- A search prompt (query)
- A set of retrieved documents

Your task is to generate a support context — a concise, well-structured summary that captures
all key facts relevant to answering the user question.

Important Instructions:
- Do not answer the question directly.
- Use only the retrieved documents.
- Do not add external knowledge.

Focus Aspect: <ASPECT>
<ASPECT EXPLANATION>

User Question:
<question>

Search Query:
<query>

Retrieved Documents:

[Doc 1] <doc_1>
[Doc 2] <doc_2>
[Doc 3] <doc_3>
\end{tabularx}

Please write the support context.
    \end{tcblisting}
    \end{minipage}
    \caption{Prompt Template used for generating the summary from GPT-4o-mini.}
    \label{fig:summary_template}
\end{figure*}

%% file: figures/case_study.tex
\begin{figure*}[h!] % Use 'figure' for a single-column float
    \centering % Center the minipage within the column
    \begin{minipage}{\textwidth} % <-- IMPORTANT: Use \columnwidth
        \begin{tcblisting}{
          colback=gray!5,
          colframe=blue!70!black,
          listing only,
          left=1mm, right=1mm, top=1mm, bottom=1mm,
          title=Summarization Output
        }
User Question: "The A Word series 2 how many episodes?"

Interpretation: The user asks a specific, factual question about episode count.

Retrieved Documents (1,821 characters)

The retriever returned three documents containing:

[Doc 1] - "The A Word"
- Information about Series 2 DVD release (December 18, 2017)
- Irrelevant: Viewer reactions on social media
- Irrelevant: Praise for acting quality
- Irrelevant: DVD bonus features details

[Doc 2] - "The A Word"
- Information about Series 2 premiere date (November 7, 2017)
- Irrelevant: Background about Israeli series "Yellow Peppers"
- Irrelevant: Plot details about five-year-old Joe Hughes
- Irrelevant: Communication problems and autism themes

[Doc 3] - "The L Word (season 2)"
- Completely irrelevant: About a different TV show entirely
- Irrelevant: Plot details about Tina and Bette
- Irrelevant: Character's alcohol abuse and job problems
- Irrelevant: Episode 12: "L'Chaim"

Analysis:
- Total retrieved content: 1,821 characters
- Estimated irrelevant content: ~70\% (plot details, unrelated shows, DVD extras, viewer reactions)
- Relevant information: Only 2--3 sentences across all documents

Summarizer Output (339 characters - 81\% reduction)

\begin{quote}
"The second series of 'The A Word' consists of six episodes. It premiered in the UK on November 7, 2017, following the commissioning announcement by the BBC on May 26, 2016. The series continues to explore the challenges faced by a family as they navigate the complexities of autism, focusing on the experiences of five-year-old Joe Hughes."
\end{quote}

What the summarizer filtered out:
- Removed: Different TV show ("The L Word")
- Removed: Social media reactions
- Removed: DVD release details and bonus features
- Removed: Irrelevant plot details from Doc 3

What the summarizer kept:
- Direct answer: "six episodes"
- Premiere date and context
- Brief series description for completeness
    \end{tcblisting}
    \end{minipage}
    \caption{Case study on the summarization output.}
    \label{fig:case_studies}
\end{figure*}

\begin{figure*}[h!] % Use 'figure' for a single-column float
    \centering % Center the minipage within the column
    \begin{minipage}{\textwidth} % <-- IMPORTANT: Use \columnwidth
        \begin{tcblisting}{
          colback=gray!5,
          colframe=blue!70!black,
          listing only,
          left=1mm, right=1mm, top=1mm, bottom=1mm,
          title=Notable Failure Patterns
        }
Case Study 1: Named Entity

Question: Who was named African footballer of the year 2014?

Ground Truth: Yaya Tour\'e

Original Documents: ``...Nigeria's Nwankwo Kanu and Jay-Jay Okocha, as well as the Ivory Coast's Yaya Touré and Egypt's Mohamed Salah are the only players to win the award more than once...''

Summary Generated: The retrieved documents do not provide specific information regarding the winner of the African Footballer of the Year for 2014. They primarily discuss the history and background of the BBC African Footballer of the Year award...

Case Study 2: Temporal Information

Question: When is the next scandal episode coming out?

Ground Truth: February 1, 2018

Original Documents: "...Find the Answer is the 54th single by Japanese boy band Arashi. It was released on February 21, 2018... used as the theme song for television drama 99.9 Keiji Senmon Bengoshi — SEASON II, which was first broadcast on January 14, 2018. [Other documents discuss 'The Quiz Show Scandal', a 2010 South Korean film, and generic question-answering systems.]"

Summary Generated: "The retrieved documents do not contain relevant information regarding the release date of the next episode of`Scandal.' They primarily discuss unrelated topics, such as a Japanese boy band's song, a collaborative question and answer platform in China, and a South Korean film."

Case Study 3: Compound Information

Question: What does HP mean in war and order?

Ground Truth: hit points or health points

Original Documents: "...HP Autonomy allows for search and processing of text... HP Information Analytics solutions help organizations extract meaning from all forms of human information...[Documents discuss HP (Hewlett-Packard) company technology, not gaming terminology]"

Summary Generated: "The term `HP' in the context of `War and Order' does not have a specific definition provided in the retrieved documents. However, the documents discuss various topics related to HP, including its software and technology... HP Autonomy... HP Information Management Software... Overall, while the documents provide insights into HP's technological capabilities, they do not clarify what `HP' specifically means in the context of `War and Order.'"
    \end{tcblisting}
    \end{minipage}
    \caption{Case study on the failure patterns.}
    \label{fig:case_studies}
\end{figure*}

%% file: neurips_2024.bbl
\begin{thebibliography}{72}
\providecommand{\natexlab}[1]{#1}
\providecommand{\url}[1]{\texttt{#1}}
\expandafter\ifx\csname urlstyle\endcsname\relax
  \providecommand{\doi}[1]{doi: #1}\else
  \providecommand{\doi}{doi: \begingroup \urlstyle{rm}\Url}\fi

\bibitem[Ahn et~al.(2024)Ahn, Verma, Lou, Liu, Zhang, and Yin]{ahn2024rejectsample}
Janice Ahn, Rishu Verma, Renze Lou, Di~Liu, Rui Zhang, and Wenpeng Yin.
\newblock Large language models for mathematical reasoning: Progresses and challenges.
\newblock In Neele Falk, Sara Papi, and Mike Zhang, editors, \emph{Proceedings of the 18th Conference of the European Chapter of the Association for Computational Linguistics: Student Research Workshop}, pages 225--237, St. Julian{'}s, Malta, March 2024. Association for Computational Linguistics.
\newblock \doi{10.18653/v1/2024.eacl-srw.17}.
\newblock URL \url{https://aclanthology.org/2024.eacl-srw.17/}.

\bibitem[Asai et~al.(2024)Asai, Wu, Wang, Sil, and Hajishirzi]{asai2024self}
Akari Asai, Zeqiu Wu, Yizhong Wang, Avi Sil, and Hannaneh Hajishirzi.
\newblock Self-rag: Learning to retrieve, generate, and critique through self-reflection.
\newblock In \emph{International conference on learning representations}, volume 2024, pages 9112--9141, 2024.

\bibitem[Bajaj et~al.(2016)Bajaj, Campos, Craswell, Deng, Gao, Liu, Majumder, McNamara, Mitra, Nguyen, et~al.]{bajaj2016msmarco}
Payal Bajaj, Daniel Campos, Nick Craswell, Li~Deng, Jianfeng Gao, Xiaodong Liu, Rangan Majumder, Andrew McNamara, Bhaskar Mitra, Tri Nguyen, et~al.
\newblock Ms marco: A human generated machine reading comprehension dataset.
\newblock \emph{arXiv preprint arXiv:1611.09268}, 2016.

\bibitem[Baker et~al.(2025)Baker, Huizinga, Gao, Dou, Guan, Madry, Zaremba, Pachocki, and Farhi]{baker2025monitoring}
Bowen Baker, Joost Huizinga, Leo Gao, Zehao Dou, Melody~Y. Guan, Aleksander Madry, Wojciech Zaremba, Jakub Pachocki, and David Farhi.
\newblock Monitoring reasoning models for misbehavior and the risks of promoting obfuscation, 2025.
\newblock URL \url{https://arxiv.org/abs/2503.11926}.

\bibitem[Cao et~al.(2025)Cao, Hegde, Li, Griggs, Liu, Tang, Pan, Wang, Malik, Neubig, Hakhamaneshi, Liaw, Moritz, Zaharia, Gonzalez, and Stoica]{cao2025skyrl}
Shiyi Cao, Sumanth Hegde, Dacheng Li, Tyler Griggs, Shu Liu, Eric Tang, Jiayi Pan, Xingyao Wang, Akshay Malik, Graham Neubig, Kourosh Hakhamaneshi, Richard Liaw, Philipp Moritz, Matei Zaharia, Joseph~E. Gonzalez, and Ion Stoica.
\newblock Skyrl-v0: Train real-world long-horizon agents via reinforcement learning, 2025.

\bibitem[Chang et~al.(2025)Chang, Jiang, Rakesh, Pan, Yeh, Wang, Hu, Xu, Zheng, Das, and Zou]{chang-etal-2025-main}
Chia-Yuan Chang, Zhimeng Jiang, Vineeth Rakesh, Menghai Pan, Chin-Chia~Michael Yeh, Guanchu Wang, Mingzhi Hu, Zhichao Xu, Yan Zheng, Mahashweta Das, and Na~Zou.
\newblock {MAIN}-{RAG}: Multi-agent filtering retrieval-augmented generation.
\newblock In Wanxiang Che, Joyce Nabende, Ekaterina Shutova, and Mohammad~Taher Pilehvar, editors, \emph{Proceedings of the 63rd Annual Meeting of the Association for Computational Linguistics (Volume 1: Long Papers)}, pages 2607--2622, Vienna, Austria, July 2025. Association for Computational Linguistics.
\newblock ISBN 979-8-89176-251-0.
\newblock \doi{10.18653/v1/2025.acl-long.131}.
\newblock URL \url{https://aclanthology.org/2025.acl-long.131/}.

\bibitem[Chen et~al.(2025)Chen, Sun, Li, Sun, Zhou, Zhu, Wang, Pan, Zhang, Chen, et~al.]{chen2025learningtoreason}
Mingyang Chen, Linzhuang Sun, Tianpeng Li, Haoze Sun, Yijie Zhou, Chenzheng Zhu, Haofen Wang, Jeff~Z Pan, Wen Zhang, Huajun Chen, et~al.
\newblock Learning to reason with search for llms via reinforcement learning.
\newblock \emph{arXiv preprint arXiv:2503.19470}, 2025.

\bibitem[Chevalier et~al.(2023)Chevalier, Wettig, Ajith, and Chen]{chevalier-etal-2023-adapting}
Alexis Chevalier, Alexander Wettig, Anirudh Ajith, and Danqi Chen.
\newblock Adapting language models to compress contexts.
\newblock In Houda Bouamor, Juan Pino, and Kalika Bali, editors, \emph{Proceedings of the 2023 Conference on Empirical Methods in Natural Language Processing}, pages 3829--3846, Singapore, December 2023. Association for Computational Linguistics.
\newblock \doi{10.18653/v1/2023.emnlp-main.232}.
\newblock URL \url{https://aclanthology.org/2023.emnlp-main.232/}.

\bibitem[Fabbri et~al.(2021)Fabbri, Kry{\'s}ci{\'n}ski, McCann, Xiong, Socher, and Radev]{fabbri2021summeval}
Alexander~R. Fabbri, Wojciech Kry{\'s}ci{\'n}ski, Bryan McCann, Caiming Xiong, Richard Socher, and Dragomir Radev.
\newblock {S}umm{E}val: Re-evaluating summarization evaluation.
\newblock \emph{Transactions of the Association for Computational Linguistics}, 9:\penalty0 391--409, 2021.
\newblock \doi{10.1162/tacl_a_00373}.
\newblock URL \url{https://aclanthology.org/2021.tacl-1.24/}.

\bibitem[Fang et~al.(2024)Fang, Bai, Ni, Yang, Chen, and Xu]{fang-etal-2024-noiserag}
Feiteng Fang, Yuelin Bai, Shiwen Ni, Min Yang, Xiaojun Chen, and Ruifeng Xu.
\newblock Enhancing noise robustness of retrieval-augmented language models with adaptive adversarial training.
\newblock In Lun-Wei Ku, Andre Martins, and Vivek Srikumar, editors, \emph{Proceedings of the 62nd Annual Meeting of the Association for Computational Linguistics (Volume 1: Long Papers)}, pages 10028--10039, Bangkok, Thailand, August 2024. Association for Computational Linguistics.
\newblock \doi{10.18653/v1/2024.acl-long.540}.
\newblock URL \url{https://aclanthology.org/2024.acl-long.540/}.

\bibitem[Feng et~al.(2025)Feng, Xue, Liu, and An]{feng2025groupingroup}
Lang Feng, Zhenghai Xue, Tingcong Liu, and Bo~An.
\newblock Group-in-group policy optimization for llm agent training.
\newblock \emph{arXiv preprint arXiv:2505.10978}, 2025.

\bibitem[Gao et~al.(2023)Gao, Xiong, Gao, Jia, Pan, Bi, Dai, Sun, Wang, and Wang]{gao2023retrievalaugmentedgeneration}
Yunfan Gao, Yun Xiong, Xinyu Gao, Kangxiang Jia, Jinliu Pan, Yuxi Bi, Yixin Dai, Jiawei Sun, Haofen Wang, and Haofen Wang.
\newblock Retrieval-augmented generation for large language models: A survey.
\newblock \emph{arXiv preprint arXiv:2312.10997}, 2\penalty0 (1), 2023.

\bibitem[Ge et~al.(2024)Ge, Jing, Wang, Wang, Chen, and Wei]{ge2024incontextautoencoder}
Tao Ge, Hu~Jing, Lei Wang, Xun Wang, Si-Qing Chen, and Furu Wei.
\newblock In-context autoencoder for context compression in a large language model.
\newblock In \emph{The Twelfth International Conference on Learning Representations}, 2024.
\newblock URL \url{https://openreview.net/forum?id=uREj4ZuGJE}.

\bibitem[Guo et~al.(2024)Guo, Zhu, Yang, Xie, Dong, Zhang, Chen, Bi, Wu, Li, et~al.]{guo2024deepseek}
Daya Guo, Qihao Zhu, Dejian Yang, Zhenda Xie, Kai Dong, Wentao Zhang, Guanting Chen, Xiao Bi, Yu~Wu, YK~Li, et~al.
\newblock Deepseek-coder: When the large language model meets programming--the rise of code intelligence.
\newblock \emph{arXiv preprint arXiv:2401.14196}, 2024.

\bibitem[Guo et~al.(2025)Guo, Yang, Zhang, Song, Zhang, Xu, Zhu, Ma, Wang, Bi, et~al.]{guo2025deepseek}
Daya Guo, Dejian Yang, Haowei Zhang, Junxiao Song, Ruoyu Zhang, Runxin Xu, Qihao Zhu, Shirong Ma, Peiyi Wang, Xiao Bi, et~al.
\newblock Deepseek-r1: Incentivizing reasoning capability in llms via reinforcement learning.
\newblock \emph{arXiv preprint arXiv:2501.12948}, 2025.

\bibitem[Ho et~al.(2020)Ho, Duong~Nguyen, Sugawara, and Aizawa]{ho-etal-2020-constructing}
Xanh Ho, Anh-Khoa Duong~Nguyen, Saku Sugawara, and Akiko Aizawa.
\newblock Constructing a multi-hop {QA} dataset for comprehensive evaluation of reasoning steps.
\newblock In Donia Scott, Nuria Bel, and Chengqing Zong, editors, \emph{Proceedings of the 28th International Conference on Computational Linguistics}, pages 6609--6625, Barcelona, Spain (Online), December 2020. International Committee on Computational Linguistics.
\newblock \doi{10.18653/v1/2020.coling-main.580}.
\newblock URL \url{https://aclanthology.org/2020.coling-main.580/}.

\bibitem[Huang et~al.(2025)Huang, Yu, Ma, Zhong, Feng, Wang, Chen, Peng, Feng, Qin, and Liu]{Huang2025SurveyOnHallucination}
Lei Huang, Weijiang Yu, Weitao Ma, Weihong Zhong, Zhangyin Feng, Haotian Wang, Qianglong Chen, Weihua Peng, Xiaocheng Feng, Bing Qin, and Ting Liu.
\newblock A survey on hallucination in large language models: Principles, taxonomy, challenges, and open questions.
\newblock \emph{{ACM} Trans. Inf. Syst.}, 2025.

\bibitem[Hurst et~al.(2024)Hurst, Lerer, Goucher, Perelman, Ramesh, Clark, Ostrow, Welihinda, Hayes, Radford, et~al.]{hurst2024gpt4osystemcard}
Aaron Hurst, Adam Lerer, Adam~P Goucher, Adam Perelman, Aditya Ramesh, Aidan Clark, AJ~Ostrow, Akila Welihinda, Alan Hayes, Alec Radford, et~al.
\newblock Gpt-4o system card.
\newblock \emph{arXiv preprint arXiv:2410.21276}, 2024.

\bibitem[Jaech et~al.(2024)Jaech, Kalai, Lerer, Richardson, El-Kishky, Low, Helyar, Madry, Beutel, Carney, et~al.]{jaech2024openaio1systemcard}
Aaron Jaech, Adam Kalai, Adam Lerer, Adam Richardson, Ahmed El-Kishky, Aiden Low, Alec Helyar, Aleksander Madry, Alex Beutel, Alex Carney, et~al.
\newblock Openai o1 system card.
\newblock \emph{arXiv preprint arXiv:2412.16720}, 2024.

\bibitem[Jiang et~al.(2023)Jiang, Wu, Lin, Yang, and Qiu]{jiang-etal-2023-llmlingua}
Huiqiang Jiang, Qianhui Wu, Chin-Yew Lin, Yuqing Yang, and Lili Qiu.
\newblock {LLML}ingua: Compressing prompts for accelerated inference of large language models.
\newblock In Houda Bouamor, Juan Pino, and Kalika Bali, editors, \emph{Proceedings of the 2023 Conference on Empirical Methods in Natural Language Processing}, pages 13358--13376, Singapore, December 2023. Association for Computational Linguistics.
\newblock \doi{10.18653/v1/2023.emnlp-main.825}.
\newblock URL \url{https://aclanthology.org/2023.emnlp-main.825/}.

\bibitem[Jin et~al.(2025{\natexlab{a}})Jin, Yoon, Han, and Arik]{jin2025longcontextllmsmeetrag}
Bowen Jin, Jinsung Yoon, Jiawei Han, and Sercan~O Arik.
\newblock Long-context llms meet rag: Overcoming challenges for long inputs in rag.
\newblock In \emph{The Thirteenth International Conference on Learning Representations}, 2025{\natexlab{a}}.

\bibitem[Jin et~al.(2025{\natexlab{b}})Jin, Yoon, Kargupta, Arik, and Han]{jin2025searchr1extension}
Bowen Jin, Jinsung Yoon, Priyanka Kargupta, Sercan~O Arik, and Jiawei Han.
\newblock An empirical study on reinforcement learning for reasoning-search interleaved llm agents.
\newblock \emph{arXiv preprint arXiv:2505.15117}, 2025{\natexlab{b}}.

\bibitem[Jin et~al.(2025{\natexlab{c}})Jin, Zeng, Yue, Yoon, Arik, Wang, Zamani, and Han]{jin2025searchr1}
Bowen Jin, Hansi Zeng, Zhenrui Yue, Jinsung Yoon, Sercan~O Arik, Dong Wang, Hamed Zamani, and Jiawei Han.
\newblock Search-r1: Training {LLM}s to reason and leverage search engines with reinforcement learning.
\newblock In \emph{Second Conference on Language Modeling}, 2025{\natexlab{c}}.
\newblock URL \url{https://openreview.net/forum?id=Rwhi91ideu}.

\bibitem[Jin et~al.(2024)Jin, Zhu, Yang, Zhang, and Dou]{jin2024FlashRAG}
Jiajie Jin, Yutao Zhu, Xinyu Yang, Chenghao Zhang, and Zhicheng Dou.
\newblock Flashrag: A modular toolkit for efficient retrieval-augmented generation research.
\newblock \emph{CoRR}, abs/2405.13576, 2024.
\newblock URL \url{https://arxiv.org/abs/2405.13576}.

\bibitem[Joshi et~al.(2017)Joshi, Choi, Weld, and Zettlemoyer]{joshi2017triviaqa}
Mandar Joshi, Eunsol Choi, Daniel Weld, and Luke Zettlemoyer.
\newblock {T}rivia{QA}: A large scale distantly supervised challenge dataset for reading comprehension.
\newblock In Regina Barzilay and Min-Yen Kan, editors, \emph{Proceedings of the 55th Annual Meeting of the Association for Computational Linguistics (Volume 1: Long Papers)}, pages 1601--1611, Vancouver, Canada, July 2017. Association for Computational Linguistics.
\newblock \doi{10.18653/v1/P17-1147}.
\newblock URL \url{https://aclanthology.org/P17-1147/}.

\bibitem[Kim and Rush(2016)]{kim-rush-2016-sequence}
Yoon Kim and Alexander~M. Rush.
\newblock Sequence-level knowledge distillation.
\newblock In Jian Su, Kevin Duh, and Xavier Carreras, editors, \emph{Proceedings of the 2016 Conference on Empirical Methods in Natural Language Processing}, pages 1317--1327, Austin, Texas, November 2016. Association for Computational Linguistics.
\newblock \doi{10.18653/v1/D16-1139}.
\newblock URL \url{https://aclanthology.org/D16-1139/}.

\bibitem[Korbak et~al.(2025)Korbak, Balesni, Barnes, Bengio, Benton, Bloom, Chen, Cooney, Dafoe, Dragan, et~al.]{korbak2025chain}
Tomek Korbak, Mikita Balesni, Elizabeth Barnes, Yoshua Bengio, Joe Benton, Joseph Bloom, Mark Chen, Alan Cooney, Allan Dafoe, Anca Dragan, et~al.
\newblock Chain of thought monitorability: A new and fragile opportunity for ai safety.
\newblock \emph{arXiv preprint arXiv:2507.11473}, 2025.

\bibitem[Kryscinski et~al.(2020)Kryscinski, McCann, Xiong, and Socher]{kryscinski2020evaluatingthefactualconsistency}
Wojciech Kryscinski, Bryan McCann, Caiming Xiong, and Richard Socher.
\newblock Evaluating the factual consistency of abstractive text summarization.
\newblock In Bonnie Webber, Trevor Cohn, Yulan He, and Yang Liu, editors, \emph{Proceedings of the 2020 Conference on Empirical Methods in Natural Language Processing (EMNLP)}, pages 9332--9346, Online, November 2020. Association for Computational Linguistics.
\newblock \doi{10.18653/v1/2020.emnlp-main.750}.
\newblock URL \url{https://aclanthology.org/2020.emnlp-main.750/}.

\bibitem[Kwiatkowski et~al.(2019)Kwiatkowski, Palomaki, Redfield, Collins, Parikh, Alberti, Epstein, Polosukhin, Devlin, Lee, et~al.]{kwiatkowski2019natural}
Tom Kwiatkowski, Jennimaria Palomaki, Olivia Redfield, Michael Collins, Ankur Parikh, Chris Alberti, Danielle Epstein, Illia Polosukhin, Jacob Devlin, Kenton Lee, et~al.
\newblock Natural questions: a benchmark for question answering research.
\newblock \emph{Transactions of the Association for Computational Linguistics}, 7:\penalty0 453--466, 2019.

\bibitem[Kwon et~al.(2023)Kwon, Li, Zhuang, Sheng, Zheng, Yu, Gonzalez, Zhang, and Stoica]{kwon2023efficient}
Woosuk Kwon, Zhuohan Li, Siyuan Zhuang, Ying Sheng, Lianmin Zheng, Cody~Hao Yu, Joseph Gonzalez, Hao Zhang, and Ion Stoica.
\newblock Efficient memory management for large language model serving with pagedattention.
\newblock In \emph{Proceedings of the 29th symposium on operating systems principles}, pages 611--626, 2023.

\bibitem[Lanham et~al.(2023)Lanham, Chen, Radhakrishnan, Steiner, Denison, Hernandez, Li, Durmus, Hubinger, Kernion, Lukovsiut.e, Nguyen, Cheng, Joseph, Schiefer, Rausch, Larson, McCandlish, Kundu, Kadavath, Yang, Henighan, Maxwell, Telleen-Lawton, Hume, Hatfield-Dodds, Kaplan, Brauner, Bowman, and Perez]{Lanham2023MeasuringFI}
Tamera Lanham, Anna Chen, Ansh Radhakrishnan, Benoit Steiner, Carson~E. Denison, Danny Hernandez, Dustin Li, Esin Durmus, Evan Hubinger, John Kernion, Kamil.e Lukovsiut.e, Karina Nguyen, Newton Cheng, Nicholas Joseph, Nicholas Schiefer, Oliver Rausch, Robin Larson, Sam McCandlish, Sandipan Kundu, Saurav Kadavath, Shannon Yang, T.~J. Henighan, Timothy~D. Maxwell, Timothy Telleen-Lawton, Tristan Hume, Zac Hatfield-Dodds, Jared Kaplan, Janina Brauner, Sam Bowman, and Ethan Perez.
\newblock Measuring faithfulness in chain-of-thought reasoning.
\newblock \emph{arXiv preprint arXiv:2307.13702}, 2023.

\bibitem[Levy et~al.(2025)Levy, Mazor, Shalmon, Hassid, and Stanovsky]{levy2025moredocuments}
Shahar Levy, Nir Mazor, Lihi Shalmon, Michael Hassid, and Gabriel Stanovsky.
\newblock More documents, same length: Isolating the challenge of multiple documents in rag.
\newblock \emph{arXiv preprint arXiv:2503.04388}, 2025.

\bibitem[Lewis et~al.(2020)Lewis, Perez, Piktus, Petroni, Karpukhin, Goyal, K\"{u}ttler, Lewis, Yih, Rockt\"{a}schel, Riedel, and Kiela]{Patrick2020rag}
Patrick Lewis, Ethan Perez, Aleksandra Piktus, Fabio Petroni, Vladimir Karpukhin, Naman Goyal, Heinrich K\"{u}ttler, Mike Lewis, Wen-tau Yih, Tim Rockt\"{a}schel, Sebastian Riedel, and Douwe Kiela.
\newblock Retrieval-augmented generation for knowledge-intensive nlp tasks.
\newblock In \emph{Advances in Neural Information Processing Systems}, pages 9459--9474. Curran Associates, Inc., 2020.

\bibitem[Li et~al.(2025)Li, Dong, Jin, Zhang, Zhou, Zhu, Zhang, and Dou]{li2025searcho1}
Xiaoxi Li, Guanting Dong, Jiajie Jin, Yuyao Zhang, Yujia Zhou, Yutao Zhu, Peitian Zhang, and Zhicheng Dou.
\newblock Search-o1: Agentic search-enhanced large reasoning models.
\newblock In Christos Christodoulopoulos, Tanmoy Chakraborty, Carolyn Rose, and Violet Peng, editors, \emph{Proceedings of the 2025 Conference on Empirical Methods in Natural Language Processing}, pages 5420--5438, Suzhou, China, November 2025. Association for Computational Linguistics.
\newblock ISBN 979-8-89176-332-6.
\newblock \doi{10.18653/v1/2025.emnlp-main.276}.
\newblock URL \url{https://aclanthology.org/2025.emnlp-main.276/}.

\bibitem[Li et~al.(2023)Li, Dong, Guerin, and Lin]{li-etal-2023-compressing}
Yucheng Li, Bo~Dong, Frank Guerin, and Chenghua Lin.
\newblock Compressing context to enhance inference efficiency of large language models.
\newblock In \emph{Proceedings of the 2023 Conference on Empirical Methods in Natural Language Processing}, pages 6342--6353, 2023.

\bibitem[Lin et~al.(2025)Lin, Wu, Xu, Liu, Tang, He, Aggarwal, Liu, Zhang, and Wang]{lin2025comprehensivesurveyreinforcementlearningbased}
Minhua Lin, Zongyu Wu, Zhichao Xu, Hui Liu, Xianfeng Tang, Qi~He, Charu Aggarwal, Hui Liu, Xiang Zhang, and Suhang Wang.
\newblock A comprehensive survey on reinforcement learning-based agentic search: Foundations, roles, optimizations, evaluations, and applications, 2025.
\newblock URL \url{https://arxiv.org/abs/2510.16724}.

\bibitem[Liu et~al.(2024)Liu, Lin, Hewitt, Paranjape, Bevilacqua, Petroni, and Liang]{liu-etal-2024-lost}
Nelson~F. Liu, Kevin Lin, John Hewitt, Ashwin Paranjape, Michele Bevilacqua, Fabio Petroni, and Percy Liang.
\newblock Lost in the middle: How language models use long contexts.
\newblock \emph{Transactions of the Association for Computational Linguistics}, 12:\penalty0 157--173, 2024.
\newblock \doi{10.1162/tacl_a_00638}.
\newblock URL \url{https://aclanthology.org/2024.tacl-1.9/}.

\bibitem[Liu et~al.(2023)Liu, Deb, Teruel, Halfaker, Radev, and Awadallah]{liu-etal-2023-improvingsummarizationfactual}
Yixin Liu, Budhaditya Deb, Milagro Teruel, Aaron Halfaker, Dragomir Radev, and Ahmed~Hassan Awadallah.
\newblock On improving summarization factual consistency from natural language feedback.
\newblock In Anna Rogers, Jordan Boyd-Graber, and Naoaki Okazaki, editors, \emph{Proceedings of the 61st Annual Meeting of the Association for Computational Linguistics (Volume 1: Long Papers)}, pages 15144--15161, Toronto, Canada, July 2023. Association for Computational Linguistics.
\newblock \doi{10.18653/v1/2023.acl-long.844}.
\newblock URL \url{https://aclanthology.org/2023.acl-long.844/}.

\bibitem[Luo et~al.(2023)Luo, Xie, and Ananiadou]{luo2023chatgptfactualinconsistencyevaluator}
Zheheng Luo, Qianqian Xie, and Sophia Ananiadou.
\newblock Chatgpt as a factual inconsistency evaluator for text summarization, 2023.
\newblock URL \url{https://arxiv.org/abs/2303.15621}.

\bibitem[Mallen et~al.(2023)Mallen, Asai, Zhong, Das, Khashabi, and Hajishirzi]{mallen-etal-2023-trust}
Alex Mallen, Akari Asai, Victor Zhong, Rajarshi Das, Daniel Khashabi, and Hannaneh Hajishirzi.
\newblock When not to trust language models: Investigating effectiveness of parametric and non-parametric memories.
\newblock In Anna Rogers, Jordan Boyd-Graber, and Naoaki Okazaki, editors, \emph{Proceedings of the 61st Annual Meeting of the Association for Computational Linguistics (Volume 1: Long Papers)}, pages 9802--9822, Toronto, Canada, July 2023. Association for Computational Linguistics.
\newblock \doi{10.18653/v1/2023.acl-long.546}.
\newblock URL \url{https://aclanthology.org/2023.acl-long.546/}.

\bibitem[Pagnoni et~al.(2021)Pagnoni, Balachandran, and Tsvetkov]{pagnoni-etal-2021-understandingfactuality}
Artidoro Pagnoni, Vidhisha Balachandran, and Yulia Tsvetkov.
\newblock Understanding factuality in abstractive summarization with {FRANK}: A benchmark for factuality metrics.
\newblock In Kristina Toutanova, Anna Rumshisky, Luke Zettlemoyer, Dilek Hakkani-Tur, Iz~Beltagy, Steven Bethard, Ryan Cotterell, Tanmoy Chakraborty, and Yichao Zhou, editors, \emph{Proceedings of the 2021 Conference of the North American Chapter of the Association for Computational Linguistics: Human Language Technologies}, pages 4812--4829, Online, June 2021. Association for Computational Linguistics.
\newblock \doi{10.18653/v1/2021.naacl-main.383}.
\newblock URL \url{https://aclanthology.org/2021.naacl-main.383/}.

\bibitem[Press et~al.(2023)Press, Zhang, Min, Schmidt, Smith, and Lewis]{press2023bamboogle}
Ofir Press, Muru Zhang, Sewon Min, Ludwig Schmidt, Noah Smith, and Mike Lewis.
\newblock Measuring and narrowing the compositionality gap in language models.
\newblock In Houda Bouamor, Juan Pino, and Kalika Bali, editors, \emph{Findings of the Association for Computational Linguistics: EMNLP 2023}, pages 5687--5711, Singapore, December 2023. Association for Computational Linguistics.
\newblock \doi{10.18653/v1/2023.findings-emnlp.378}.
\newblock URL \url{https://aclanthology.org/2023.findings-emnlp.378/}.

\bibitem[Rae et~al.(2020)Rae, Potapenko, Jayakumar, Hillier, and Lillicrap]{Rae2020Compressive}
Jack~W. Rae, Anna Potapenko, Siddhant~M. Jayakumar, Chloe Hillier, and Timothy~P. Lillicrap.
\newblock Compressive transformers for long-range sequence modelling.
\newblock In \emph{International Conference on Learning Representations}, 2020.
\newblock URL \url{https://openreview.net/forum?id=SylKikSYDH}.

\bibitem[Ryu et~al.(2024)Ryu, Do, Kim, Lee, and Ok]{ryu-etal-2024-multidimensional}
Sangwon Ryu, Heejin Do, Yunsu Kim, Gary Lee, and Jungseul Ok.
\newblock Multi-dimensional optimization for text summarization via reinforcement learning.
\newblock In Lun-Wei Ku, Andre Martins, and Vivek Srikumar, editors, \emph{Proceedings of the 62nd Annual Meeting of the Association for Computational Linguistics (Volume 1: Long Papers)}, pages 5858--5871, Bangkok, Thailand, August 2024. Association for Computational Linguistics.
\newblock \doi{10.18653/v1/2024.acl-long.319}.
\newblock URL \url{https://aclanthology.org/2024.acl-long.319/}.

\bibitem[Schick et~al.(2023)Schick, Dwivedi-Yu, Dess{\`\i}, Raileanu, Lomeli, Hambro, Zettlemoyer, Cancedda, and Scialom]{schick2023toolformer}
Timo Schick, Jane Dwivedi-Yu, Roberto Dess{\`\i}, Roberta Raileanu, Maria Lomeli, Eric Hambro, Luke Zettlemoyer, Nicola Cancedda, and Thomas Scialom.
\newblock Toolformer: Language models can teach themselves to use tools.
\newblock \emph{Advances in Neural Information Processing Systems}, 36:\penalty0 68539--68551, 2023.

\bibitem[Schulman et~al.(2017)Schulman, Wolski, Dhariwal, Radford, and Klimov]{schulman2017proximal}
John Schulman, Filip Wolski, Prafulla Dhariwal, Alec Radford, and Oleg Klimov.
\newblock Proximal policy optimization algorithms.
\newblock \emph{arXiv preprint arXiv:1707.06347}, 2017.

\bibitem[Shao et~al.(2024)Shao, Wang, Zhu, Xu, Song, Bi, Zhang, Zhang, Li, Wu, et~al.]{shao2024deepseekmath}
Zhihong Shao, Peiyi Wang, Qihao Zhu, Runxin Xu, Junxiao Song, Xiao Bi, Haowei Zhang, Mingchuan Zhang, YK~Li, Y~Wu, et~al.
\newblock Deepseekmath: Pushing the limits of mathematical reasoning in open language models.
\newblock \emph{arXiv preprint arXiv:2402.03300}, 2024.

\bibitem[Sheng et~al.(2025)Sheng, Zhang, Ye, Wu, Zhang, Zhang, Peng, Lin, and Wu]{sheng2024hybridflow}
Guangming Sheng, Chi Zhang, Zilingfeng Ye, Xibin Wu, Wang Zhang, Ru~Zhang, Yanghua Peng, Haibin Lin, and Chuan Wu.
\newblock Hybridflow: A flexible and efficient rlhf framework.
\newblock In \emph{Proceedings of the Twentieth European Conference on Computer Systems}, EuroSys '25, page 1279–1297, New York, NY, USA, 2025. Association for Computing Machinery.
\newblock ISBN 9798400711961.
\newblock \doi{10.1145/3689031.3696075}.
\newblock URL \url{https://doi.org/10.1145/3689031.3696075}.

\bibitem[Song et~al.(2025)Song, Jiang, Min, Chen, Chen, Zhao, Fang, and Wen]{song2025r1searcher}
Huatong Song, Jinhao Jiang, Yingqian Min, Jie Chen, Zhipeng Chen, Wayne~Xin Zhao, Lei Fang, and Ji-Rong Wen.
\newblock R1-searcher: Incentivizing the search capability in llms via reinforcement learning.
\newblock \emph{arXiv preprint arXiv:2503.05592}, 2025.

\bibitem[Sun et~al.(2025)Sun, Zhong, Zhou, and Han]{sun2025dynamicrag}
Jiashuo Sun, Xianrui Zhong, Sizhe Zhou, and Jiawei Han.
\newblock Dynamicrag: Leveraging outputs of large language model as feedback for dynamic reranking in retrieval-augmented generation.
\newblock \emph{arXiv preprint arXiv:2505.07233}, 2025.

\bibitem[Sutton et~al.(1998)Sutton, Barto, et~al.]{sutton1998reinforcement}
Richard~S Sutton, Andrew~G Barto, et~al.
\newblock \emph{Reinforcement learning: An introduction}.
\newblock MIT press Cambridge, 1998.

\bibitem[Trivedi et~al.(2022)Trivedi, Balasubramanian, Khot, and Sabharwal]{trivedi2022musique}
Harsh Trivedi, Niranjan Balasubramanian, Tushar Khot, and Ashish Sabharwal.
\newblock Musique: Multihop questions via single-hop question composition.
\newblock \emph{Transactions of the Association for Computational Linguistics}, 10:\penalty0 539--554, 2022.
\newblock \doi{10.1162/tacl_a_00475}.
\newblock URL \url{https://aclanthology.org/2022.tacl-1.31/}.

\bibitem[Trivedi et~al.(2023)Trivedi, Balasubramanian, Khot, and Sabharwal]{trivedi2023ircot}
Harsh Trivedi, Niranjan Balasubramanian, Tushar Khot, and Ashish Sabharwal.
\newblock Interleaving retrieval with chain-of-thought reasoning for knowledge-intensive multi-step questions.
\newblock In \emph{Proceedings of the 61st Annual Meeting of the Association for Computational Linguistics (Volume 1: Long Papers)}, pages 10014--10037, Toronto, Canada, 2023. Association for Computational Linguistics.

\bibitem[Wang et~al.(2022)Wang, Yang, Huang, Jiao, Yang, Jiang, Majumder, and Wei]{wang2022text}
Liang Wang, Nan Yang, Xiaolong Huang, Binxing Jiao, Linjun Yang, Daxin Jiang, Rangan Majumder, and Furu Wei.
\newblock Text embeddings by weakly-supervised contrastive pre-training.
\newblock \emph{arXiv preprint arXiv:2212.03533}, 2022.

\bibitem[Wang et~al.(2023)Wang, Araki, Jiang, Parvez, and Neubig]{wang-etal-2024-learning-to-filter}
Zhiruo Wang, Jun Araki, Zhengbao Jiang, Md~Rizwan Parvez, and Graham Neubig.
\newblock Learning to filter context for retrieval-augmented generation, 2023.
\newblock URL \url{https://arxiv.org/abs/2311.08377}.

\bibitem[Wei et~al.(2022)Wei, Wang, Schuurmans, Bosma, Xia, Chi, Le, Zhou, et~al.]{wei2022chain}
Jason Wei, Xuezhi Wang, Dale Schuurmans, Maarten Bosma, Fei Xia, Ed~Chi, Quoc~V Le, Denny Zhou, et~al.
\newblock Chain-of-thought prompting elicits reasoning in large language models.
\newblock \emph{Advances in neural information processing systems}, 35:\penalty0 24824--24837, 2022.

\bibitem[Wu et~al.(2025)Wu, Li, Zhao, Zhang, Ou, Yin, Zhang, Jiang, Xie, Huang, et~al.]{wu2025resum}
Xixi Wu, Kuan Li, Yida Zhao, Liwen Zhang, Litu Ou, Huifeng Yin, Zhongwang Zhang, Yong Jiang, Pengjun Xie, Fei Huang, et~al.
\newblock Resum: Unlocking long-horizon search intelligence via context summarization.
\newblock \emph{arXiv preprint arXiv:2509.13313}, 2025.

\bibitem[Xu et~al.(2024)Xu, Shi, and Choi]{xu2024recomp}
Fangyuan Xu, Weijia Shi, and Eunsol Choi.
\newblock {RECOMP}: Improving retrieval-augmented {LM}s with context compression and selective augmentation.
\newblock In \emph{The Twelfth International Conference on Learning Representations}, 2024.
\newblock URL \url{https://openreview.net/forum?id=mlJLVigNHp}.

\bibitem[Xu and Lapata(2020)]{xu2020coarse}
Yumo Xu and Mirella Lapata.
\newblock Coarse-to-fine query focused multi-document summarization.
\newblock In Bonnie Webber, Trevor Cohn, Yulan He, and Yang Liu, editors, \emph{Proceedings of the 2020 Conference on Empirical Methods in Natural Language Processing (EMNLP)}, pages 3632--3645, Online, November 2020. Association for Computational Linguistics.
\newblock \doi{10.18653/v1/2020.emnlp-main.296}.
\newblock URL \url{https://aclanthology.org/2020.emnlp-main.296/}.

\bibitem[Xu et~al.(2025{\natexlab{a}})Xu, Feng, Tian, Ding, and Cheong]{xu-etal-2025-csplade}
Zhichao Xu, Aosong Feng, Yijun Tian, Haibo Ding, and Lin~Lee Cheong.
\newblock {CSPLADE}: Learned sparse retrieval with causal language models.
\newblock In Kentaro Inui, Sakriani Sakti, Haofen Wang, Derek~F. Wong, Pushpak Bhattacharyya, Biplab Banerjee, Asif Ekbal, Tanmoy Chakraborty, and Dhirendra~Pratap Singh, editors, \emph{Proceedings of the 14th International Joint Conference on Natural Language Processing and the 4th Conference of the Asia-Pacific Chapter of the Association for Computational Linguistics}, pages 99--114, Mumbai, India, December 2025{\natexlab{a}}. The Asian Federation of Natural Language Processing and The Association for Computational Linguistics.
\newblock ISBN 979-8-89176-298-5.
\newblock \doi{10.18653/v1/2025.ijcnlp-long.7}.
\newblock URL \url{https://aclanthology.org/2025.ijcnlp-long.7/}.

\bibitem[Xu et~al.(2025{\natexlab{b}})Xu, Huang, Zhuang, and Srikumar]{xu-etal-2025-distillation}
Zhichao Xu, Zhiqi Huang, Shengyao Zhuang, and Vivek Srikumar.
\newblock Distillation versus contrastive learning: How to train your rerankers.
\newblock In Kentaro Inui, Sakriani Sakti, Haofen Wang, Derek~F. Wong, Pushpak Bhattacharyya, Biplab Banerjee, Asif Ekbal, Tanmoy Chakraborty, and Dhirendra~Pratap Singh, editors, \emph{Proceedings of the 14th International Joint Conference on Natural Language Processing and the 4th Conference of the Asia-Pacific Chapter of the Association for Computational Linguistics}, pages 564--578, Mumbai, India, December 2025{\natexlab{b}}. The Asian Federation of Natural Language Processing and The Association for Computational Linguistics.
\newblock ISBN 979-8-89176-303-6.
\newblock \doi{10.18653/v1/2025.findings-ijcnlp.33}.
\newblock URL \url{https://aclanthology.org/2025.findings-ijcnlp.33/}.

\bibitem[Xu et~al.(2025{\natexlab{c}})Xu, Yan, Gupta, and Srikumar]{xu-etal-2025-state}
Zhichao Xu, Jinghua Yan, Ashim Gupta, and Vivek Srikumar.
\newblock State space models are strong text rerankers.
\newblock In Vaibhav Adlakha, Alexandra Chronopoulou, Xiang~Lorraine Li, Bodhisattwa~Prasad Majumder, Freda Shi, and Giorgos Vernikos, editors, \emph{Proceedings of the 10th Workshop on Representation Learning for NLP (RepL4NLP-2025)}, pages 152--169, Albuquerque, NM, May 2025{\natexlab{c}}. Association for Computational Linguistics.
\newblock ISBN 979-8-89176-245-9.
\newblock \doi{10.18653/v1/2025.repl4nlp-1.12}.
\newblock URL \url{https://aclanthology.org/2025.repl4nlp-1.12/}.

\bibitem[Xu et~al.(2026{\natexlab{a}})Xu, Mo, Huang, Zhang, Yu, Phillips, Lin, and Srikumar]{xu2025surveyofmodelarchitectures}
Zhichao Xu, Fengran Mo, Zhiqi Huang, Crystina Zhang, Puxuan Yu, Bei~Wang Phillips, Jimmy Lin, and Vivek Srikumar.
\newblock A survey of model architectures in information retrieval.
\newblock \emph{Transactions on Machine Learning Research}, 2026{\natexlab{a}}.
\newblock ISSN 2835-8856.
\newblock URL \url{https://openreview.net/forum?id=xAIbTbHRrX}.
\newblock Survey Certification.

\bibitem[Xu et~al.(2026{\natexlab{b}})Xu, Wu, Zhou, Feng, Zhou, Woo, Ramnath, Tian, Qi, Qiu, Cheong, and Ding]{xu2026beyondcorrectness}
Zhichao Xu, Zongyu Wu, Yun Zhou, Aosong Feng, Kang Zhou, Sangmin Woo, Kiran Ramnath, Yijun Tian, Xuan Qi, Weikang Qiu, Lin~Lee Cheong, and Haibo Ding.
\newblock Beyond correctness: Rewarding faithful reasoning in retrieval-augmented generation.
\newblock \emph{Transactions on Machine Learning Research}, 2026{\natexlab{b}}.
\newblock ISSN 2835-8856.
\newblock URL \url{https://openreview.net/forum?id=mZ0gGlXelF}.

\bibitem[Xu et~al.(2026{\natexlab{c}})Xu, Zhuang, Ma, Chen, Tian, Mo, Li, Cao, and Srikumar]{xu2026rethinkingonpolicyoptimizationquery}
Zhichao Xu, Shengyao Zhuang, Xueguang Ma, Bingsen Chen, Yijun Tian, Fengran Mo, Tao Li, Jie Cao, and Vivek Srikumar.
\newblock Rethinking on-policy optimization for query augmentation, 2026{\natexlab{c}}.
\newblock URL \url{https://arxiv.org/abs/2510.17139}.

\bibitem[Yang et~al.(2024)Yang, Yang, Zhang, Hui, Zheng, Yu, Li, Liu, Huang, Wei, Lin, Yang, Tu, Zhang, Yang, Yang, Zhou, Lin, Dang, Lu, Bao, Yang, Yu, Li, Xue, Zhang, Zhu, Men, Lin, Li, Xia, Ren, Ren, Fan, Su, Zhang, Wan, Liu, Cui, Zhang, and Qiu]{yang2024qwen25}
An~Yang, Baosong Yang, Beichen Zhang, Binyuan Hui, Bo~Zheng, Bowen Yu, Chengyuan Li, Dayiheng Liu, Fei Huang, Haoran Wei, Huan Lin, Jian Yang, Jianhong Tu, Jianwei Zhang, Jianxin Yang, Jiaxi Yang, Jingren Zhou, Junyang Lin, Kai Dang, Keming Lu, Keqin Bao, Kexin Yang, Le~Yu, Mei Li, Mingfeng Xue, Pei Zhang, Qin Zhu, Rui Men, Runji Lin, Tianhao Li, Tingyu Xia, Xingzhang Ren, Xuancheng Ren, Yang Fan, Yang Su, Yichang Zhang, Yu~Wan, Yuqiong Liu, Zeyu Cui, Zhenru Zhang, and Zihan Qiu.
\newblock Qwen2.5 technical report.
\newblock \emph{arXiv preprint arXiv:2412.15115}, 2024.

\bibitem[Yang et~al.(2018)Yang, Qi, Zhang, Bengio, Cohen, Salakhutdinov, and Manning]{yang2018hotpotqa}
Zhilin Yang, Peng Qi, Saizheng Zhang, Yoshua Bengio, William Cohen, Ruslan Salakhutdinov, and Christopher~D. Manning.
\newblock {H}otpot{QA}: A dataset for diverse, explainable multi-hop question answering.
\newblock In Ellen Riloff, David Chiang, Julia Hockenmaier, and Jun{'}ichi Tsujii, editors, \emph{Proceedings of the 2018 Conference on Empirical Methods in Natural Language Processing}, pages 2369--2380, Brussels, Belgium, October-November 2018. Association for Computational Linguistics.
\newblock \doi{10.18653/v1/D18-1259}.
\newblock URL \url{https://aclanthology.org/D18-1259/}.

\bibitem[Yao et~al.(2023)Yao, Zhao, Yu, Du, Shafran, Narasimhan, and Cao]{yao2023react}
Shunyu Yao, Jeffrey Zhao, Dian Yu, Nan Du, Izhak Shafran, Karthik~R Narasimhan, and Yuan Cao.
\newblock React: Synergizing reasoning and acting in language models.
\newblock In \emph{The Eleventh International Conference on Learning Representations}, 2023.
\newblock URL \url{https://openreview.net/forum?id=WE_vluYUL-X}.

\bibitem[Yoran et~al.(2024)Yoran, Wolfson, Ram, and Berant]{yoran2024makingretrievalaugmented}
Ori Yoran, Tomer Wolfson, Ori Ram, and Jonathan Berant.
\newblock Making retrieval-augmented language models robust to irrelevant context.
\newblock In \emph{The Twelfth International Conference on Learning Representations}, 2024.
\newblock URL \url{https://openreview.net/forum?id=ZS4m74kZpH}.

\bibitem[Zheng et~al.(2024)Zheng, Zhang, Zhang, Ye, and Luo]{zheng-etal-2024-llamafactory}
Yaowei Zheng, Richong Zhang, Junhao Zhang, Yanhan Ye, and Zheyan Luo.
\newblock {L}lama{F}actory: Unified efficient fine-tuning of 100+ language models.
\newblock In Yixin Cao, Yang Feng, and Deyi Xiong, editors, \emph{Proceedings of the 62nd Annual Meeting of the Association for Computational Linguistics (Volume 3: System Demonstrations)}, pages 400--410, Bangkok, Thailand, August 2024. Association for Computational Linguistics.
\newblock \doi{10.18653/v1/2024.acl-demos.38}.
\newblock URL \url{https://aclanthology.org/2024.acl-demos.38/}.

\bibitem[Zheng et~al.(2025)Zheng, Fu, Hu, Cai, Ye, Lu, and Liu]{zheng2025deepresearcher}
Yuxiang Zheng, Dayuan Fu, Xiangkun Hu, Xiaojie Cai, Lyumanshan Ye, Pengrui Lu, and Pengfei Liu.
\newblock Deepresearcher: Scaling deep research via reinforcement learning in real-world environments.
\newblock \emph{arXiv preprint arXiv:2504.03160}, 2025.

\bibitem[Zhu et~al.(2021)Zhu, Hinthorn, Xu, Zeng, Zeng, Huang, and Jiang]{zhu2021enhancingfactualconsistency}
Chenguang Zhu, William Hinthorn, Ruochen Xu, Qingkai Zeng, Michael Zeng, Xuedong Huang, and Meng Jiang.
\newblock Enhancing factual consistency of abstractive summarization.
\newblock In Kristina Toutanova, Anna Rumshisky, Luke Zettlemoyer, Dilek Hakkani-Tur, Iz~Beltagy, Steven Bethard, Ryan Cotterell, Tanmoy Chakraborty, and Yichao Zhou, editors, \emph{Proceedings of the 2021 Conference of the North American Chapter of the Association for Computational Linguistics: Human Language Technologies}, pages 718--733, Online, June 2021. Association for Computational Linguistics.
\newblock \doi{10.18653/v1/2021.naacl-main.58}.
\newblock URL \url{https://aclanthology.org/2021.naacl-main.58/}.

\end{thebibliography}
